\documentclass[Afour,sageh,times]{sagej}

\usepackage{moreverb,url}
\usepackage{array}
\usepackage{multirow}
\usepackage{mdwlist}
\usepackage{epsfig} % for postscript graphics files
\usepackage{epstopdf}
\usepackage{graphics}
\usepackage{graphicx}

\newcommand\BibTeX{{\rmfamily B\kern-.05em \textsc{i\kern-.025em b}\kern-.08em
T\kern-.1667em\lower.7ex\hbox{E}\kern-.125emX}}

\begin{document}

\runninghead{Faraji et al.}

\title{Push recovery with stepping strategy based on time-projection control}

\author{Salman Faraji\affilnum{1}, Hamed Razavi\affilnum{1} and Auke J. Ijspeert\affilnum{1}}

\affiliation{\affilnum{1}EPFL, Switzerland}

\corrauth{Salman Faraji, EPFL STI IBI-STI BIOROB, Station 9, CH-1015 Lausanne, Switzerland}

\email{salman.faraji@epfl.ch}

\begin{abstract}
In this paper, we present a simple control framework for on-line push recovery with dynamic stepping properties. Due to relatively heavy legs in our robot, we need to take swing dynamics into account and thus use a linear model called 3LP which is composed of three pendulums to simulate swing and torso dynamics. Based on 3LP equations, we formulate discrete LQR controllers and use a particular time-projection method to adjust the next footstep location on-line during the motion continuously. This adjustment, which is found based on both pelvis and swing foot tracking errors, naturally takes the swing dynamics into account. Suggested adjustments are added to the Cartesian 3LP gaits and converted to joint-space trajectories through inverse kinematics. Fixed and adaptive foot lift strategies also ensure enough ground clearance in perturbed walking conditions. The proposed structure is robust, yet uses very simple state estimation and basic position tracking. We rely on the physical series elastic actuators to absorb impacts while introducing simple laws to compensate their tracking bias. Extensive experiments demonstrate the functionality of different control blocks and prove the effectiveness of time-projection in extreme push recovery scenarios. We also show self-produced and emergent walking gaits when the robot is subject to continuous dragging forces. These gaits feature dynamic walking robustness due to relatively soft springs in the ankles and avoiding any Zero Moment Point (ZMP) control in our proposed architecture. 
\end{abstract}


\maketitle

%%%%%%%%%%%%%%%%%%%%%%%%%%%%%%%%%%%%%%%%%%%%%%%%%%%%%%%%%%%%%%%%%%%%%%%%%%
%%%%%%%%%%%%%%%%%%%%%%%%%%%%%%%%%%%%%%%%%%%%%%%%%%%%%%%%%%%%%%%%%%%%%%%%%%
%%%%%%%%%%%%%%%%%%%%%%%%%%%%%%%%%%%%%%%%%%%%%%%%%%%%%%%%%%%%%%%%%%%%%%%%%%
%%%%%%%%%%%%%%%%%%%%%%%%%%%%%%%%%%%%%%%%%%%%%%%%%%%%%%%%%%%%%%%%%%%%%%%%%%
\section{Introduction}

Humanoid robots are designed to perform different locomotion and manipulation tasks like humans by exploiting similar kinematics and mass distribution properties. In this work, we aim at walking gait generation with the emphasis on push recovery properties and stepping strategy. We discuss critical ingredients needed in perturbed walking conditions together with requirements that the particular hardware platform imposes on the control design. Based on these concepts then, we introduce a very simple controller that benefits from well-established theories in previous works but implements look-up-table control laws to stabilize the robot and recover external disturbances. The proposed controller is very simple to implement, computationally very fast and yet generic with no critical parameter to tune. We will continue this section by introducing key concepts required for perturbed walking. We explain our walking model and control theories developed in previous works in the following two sections briefly. All low-level control strategies that tackle hardware limitations, including velocity limits, delays, noises, backlashes, spring deflections and ground clearance strategies, are explained in a separate section afterward. We conclude the paper then by a wide range of experiments that demonstrate the effectiveness of our method to perform dynamic walking and push recovery. 

\subsection{Compliance in Walking}

Virtual or physical leg compliance is a determinant feature of humanoid walking on uneven terrain or perturbed conditions. In the absence of severe impacts, e.g., on perfectly known terrains or in static locomotion conditions \cite{rebula2007controller}, compliance does not play an important role. In periodic flat-ground walking also, the control method can be adapted to minimize the impacts and smoothen the locomotion. Non-periodic walking conditions, however, require more compliance due to an unexpected timing of contact phase changes. Many walking controllers rely on swing leg compliance and contact force sensors to provide a phase-based or adaptive contact switch \cite{capturability2, gehring2013control, kelly2016off, lim2001balance}. Phase detection through contact force measurement helps to choose the right swing/stance controller for each leg which improves stability \cite{aoi2006stability, faber2007stochastic, collins2005bipedal}, despite a fixed timing in the desired periodic gait. A simple strategy could be early termination or polynomial extrapolation of trajectories in case of early or late phase transition respectively \cite{faraji2013compliant}.

Compliance and damping can prevent non-smooth changes in velocities over impacts \cite{park1999impedance} and make state estimation easier. Besides, adding Series Elastic Elements (SEA) in the joints help to reduce impacts and to protect the actuators. However, one requires an acceptable level of precision in state estimation and actuation. This translates to a proper spring deflection \cite{hutter2017anymal} or force gauge measurement \cite{faraji2014robust} for a closed-loop torque control. With SEA actuators, a stiff position control is also possible, but the physical springs typically determine the effective stiffness due to softer properties. Actuators with SEA elements can improve the efficiency of walking too. Depending on the spring size and actuator's isometric force generation efficiency, robots or prosthetic devices can absorb energy during impacts and rerelease it when redirecting the Center of Mass (CoM) velocity \cite{sreenath2011compliant, collins2005efficient, grimmer2016powered}. In some robots like COMAN \cite{coman} or MIT's Spring Flamingo \cite{pratt1998intuitive}, however, the SEA elements are not big enough to store a considerable amount of energy. Overall, the walking controller should not fight against SEA's functionality. In other words, the controller should demonstrate at least a passive behavior to let the springs absorb impacts or store energy smoothly \cite{sreenath2011compliant}. In this work, since our robot COMAN has small springs, we are mainly interested in transient push recovery conditions and stability rather than energetics and efficiency.

\subsection{Swing Dynamics}

Due to a small size (height of $\approx90\ cm$ and mass of $\approx30\ kg$), COMAN requires a relatively high frequency for stepping. The stepping time in a child of about the same stature (age range of $3-4$ years old, height of $105\pm2\ cm$ and mass of $17.3\pm0.7\ kg$) is about $0.45\ s$ while that in a child with about the same mass (age range of $6-7$ years old, height of $125\pm1\ cm$ and mass of $25.3\pm0.9\ kg$) is about $0.47\ s$ at self-selected speed \cite{hausdorff1999maturation}. On the one hand, COMAN is more massive than an average child of the same height. On the other hand, the leg mass in COMAN is about $22.5\%$ of the body mass while this ratio is about $17.3\%$ in human \cite{de1996adjustments}. Stepping times longer than $0.5\ s$ in COMAN lead to large lateral bounces and static motions (e.g., $1\ s$ in \cite{kryczka2015online}). Besides, unlike MABEL \cite{sreenath2011compliant} and many other biped robots, ankle actuators are distal to the knee in COMAN which further increases the leg inertia around the hip.

With a small size, yet relatively heavy legs and high inertias, the swing dynamics in COMAN has a considerable influence on CoM dynamics. Walking at a high frequency could be challenging due to actuator tracking delays, extra torques needed in the hip joints, fast ground clearance and knee motions, filter delays in state estimation and impacts in case of perturbations. However, at the same walking speed, increasing the frequency reduces the step length which makes application of linear constant CoM height models easier. Besides, SEA elements help to reduce intrinsic impacts of such a high cadence, which makes dynamic walking possible with COMAN too. This goal requires an advanced gait generation method which considers swing dynamics and internal coupling between lower limbs and the upper body. 

\subsection{Gait Generation}

In a recent work \cite{faraji2014robust}, we used the Linear Inverted Pendulum (LIP) model \cite{kajita20013d} with closed-form equations in a Model Predictive Control (MPC) framework to adjust footsteps in an online fashion. We also introduced an artificial swing trajectory which terminated at the optimized footstep location, hoping that the underlying inverse dynamics can track the template motion. The available control authority for inverse dynamics was modulating the Center of Pressure (CoP) to track the CoM and swing trajectories. Such hierarchical controller worked at a walking frequency of $2\ steps/s$ and speeds up to $0.35\ m/s$ in a perfect simulation environment. 

Tracking issues in the inverse dynamics layer (due to a limited foot size) motivated us to extend LIP with two other pendulums to model the swing leg and the torso. The new linear model, called 3LP \cite{faraji20173lp}, assumes ideal actuators in the legs to keep limb masses and the pelvis in constant height planes. It also assumes a perfect stance hip actuator to keep the torso upright. 3LP allows different stance ankle and swing hip torque profiles as inputs which can create various walking gait patterns. Thanks to inclusion of swing and torso dynamics, 3LP can produce more human-like CoM trajectories compared to LIP \cite{faraji20173lp} which facilitates tracking for the underlying inverse dynamics controller. 

In this work, we focus on the sagittal plane push recovery performance while leaving the robot bounce left and right with a fixed frequency of $2.5\ steps/s$ (step time of $0.4\ s$) to reduce lateral bounces. We assume decoupled lateral and sagittal dynamics \cite{kajita2001real, kajita2003biped} which requires smaller step lengths (or large feet to provide transverse torques) in practice. Equations of 3LP are also decoupled in lateral and sagittal directions which further supports this assumption. Due to inclusion of pelvis, 3LP is restricted to straight walking unlike the 3D LIP model we used in \cite{faraji2014robust} for steering. However, the pelvis produces a natural lateral bouncing in 3LP that does not need to enforce artificial feet separation like \cite{faraji2014robust}. 

For open-loop gait generation in the present work, we use Cartesian trajectories of 3LP and convert them to joint trajectories. Such unstable walking gait in the sagittal plane can be stabilized with a time-projecting controller \cite{faraji20173lp2}. We only focus on stabilization of sagittal plane dynamics in this work and rely on intrinsic limited stability of lateral plane bounces through a combination of waist roll \cite{hip_roll_webian_waist} and leg lift \cite{collins2001three} clearance strategies without any ZMP control \cite{zhao2008humanoid}, momentum control \cite{kuo1999stabilization}, lateral footstep adjustment \cite{kryczka2015online} and variable timing \cite{capturability2}. This combination is particularly chosen to reduce knee joint motions (motivated by limited maximum joint velocities of our particular hardware) during ground clearance and produce more stretched-knee postures \cite{hip_roll_webian_waist, ogura2003stretch}. 

\subsection{Control Approach}

Due to the fast nature of falling dynamics, the walking cycle has an unstable mode whose strength depends on the robot size and walking frequency \cite{faraji20173lp2}. A popular way of stabilizing the gait is to consider specific events (e.g., touch-down or mid-stance) and obtain Poincar\'e maps \cite{poincare} as discrete models. Classical control techniques (e.g. Linear Quadratic Regulators (LQR) \cite{ogata1995discrete}) can then stabilize the system by proper control input adjustment \cite{zaytsev2015two, kelly2015non, byl2008approximate, rummel2010stable}. The idea of phase space constant time to velocity reversal planner \cite{kim2016stabilizing} also falls in the same category where a new footstep location is calculated only once per step. Although the computations involved are CPU intensive and the update rate is relatively slow, Kim achieved 18 successful steps on a foot-less robot called Hume \cite{kim2016stabilizing}. This is considerable given transversal slippages of the point-feet, but an external motion capture system helped the state estimation in this work. Other key elements that helped Hume achieve this performance were a relatively high CoM, a high stepping frequency, and relatively light-weight legs. In \cite{faraji20173lp2}, by considering the effect of strong inter-phase disturbances, we motivated the need for an on-line adjustment of the next footstep location. Due to a relatively weak coupling of swing and CoM dynamics in slow walking conditions, adjusting the next footstep location becomes effective only in the next phase where this position becomes the new stance position which is tightly coupled to other system variables \cite{raibert1986legged, capturability}. Modulating the CoP through ankle torques can provide immediate stabilization, but is only useful in small disturbance conditions \cite{asano2004novel, coman}. 

On-line adjustment of footstep locations is widely studied in the literature to recover strong pushes \cite{faraji2014robust, feng20133d, herdt2010walking, capturability} during walking. However, the underlying simple template models (like LIP) require inverse dynamics or ZMP controllers to use ankles for a better tracking. Therefore, a combination of stepping and CoP modulation strategies is usually used \cite{faraji2014robust, kryczka2015online, feng20133d} for humanoid walking. Intuitive attack angle adjustment rules which capture the extra CoM velocity in hopping \cite{raibert1986legged} are extended to walking conditions too through capturability framework \cite{capturability}. That framework aims to find a footstep location that captures the motion and brings the CoM to rest conditions. One significant strength of the capturability framework is considering certain limitations, particularly the step length which is typically due to the use of crouched knees and the LIP model. Given such constraints, capture regions can be calculated for multiple steps as well when disturbances are too large. For walking generation, however, a proportional gain is practically needed to move the desired capture point away systematically to let the robot progress forward \cite{capturability2}. Using the 3LP model \cite{faraji20173lp} for walking generation potentially solves this issue. Besides, we use a previously-developed unified controller \cite{faraji20173lp2} in the present article which works for arbitrary walking speeds. Capturability offers an efficient start/stop state machine \cite{capturability2} which is not implemented in our work yet and considered for future improvements.

Previously, we extended the capturability idea and formulated a MPC controller that implemented a similar concept and allowed for automatic gait generation (based on LIP) inside the MPC \cite{faraji2014robust}. At each instance of time, this MPC controller simulated evolution of the measured state over a receding horizon and optimized footstep locations based on a given desired velocity. A simple arc trajectory then determined instantaneous foot accelerations which, together with the CoM accelerations (according to LIP) were converted to joint torques through inverse dynamics. The IMU placed on the pelvis also helped to regulate the torso orientation which leads to a natural falling with the given CoM accelerations. This framework was later extended by Feng \cite{feng2016robust} and applied to Atlas for push recovery and walking generation which featured robustness. The control approach we use in the present paper is very similar to the mentioned MPC controllers \cite{faraji2014robust, feng2016robust} regarding performance, but much simpler in structure.

\subsubsection{High-Level Stabilization:}
On top of the 3LP gaits, one can derive transition equations like Poincar\'e maps and use an infinite horizon discrete LQR (DLQR) controller for footstep adjustment \cite{faraji20173lp}. This architecture offers the same optimality as our MPC, but due to a discrete nature, it is sensitive to intermittent short disturbances. Therefore, we introduced a simple time-projection scheme in \cite{faraji20173lp2} which maps measured errors at any time to the previous discrete event and uses the expertise of DLQR to find on-line footstep adjustments. In the present work, we use gait generation of the 3LP model, optimality of LQR and online properties of time-projection to achieve the same performance as the MPC \cite{faraji2014robust, feng2016robust}. The new framework offers more natural gaits and faster computations because of the 3LP model and time-projection control respectively. It takes both the CoM and swing errors (positions and velocities) into account to ensures stability for a high-frequency gait and considerable swing dynamics. Compared to Raibert's law \cite{raibert1986legged} and the capturability framework \cite{capturability}, it effectively adjusts the attack angle by variable gains (on CoM and swing errors) which are found systematically through the DLQR design and time-projection. These gains are also consistent with 3LP's falling and swing dynamics. 

\subsubsection{Low-level Joint Control:}
Motivated by the specific anthropomorphic features of our COMAN robot, we use the 3LP model instead of LIP in this work which provides more natural CoM and swing trajectories and improves tracking. Although we have achieved a convincing balancing performance through pure torque control and inverse dynamics earlier \cite{faraji2015practical}, we found the pure torque control less precise in the absence of position or velocity tracking for faster-walking tasks. In the literature also, a combination of torque, velocity and position controllers is typically used to ensure compliant and precise tracking \cite{capturability2, feng2015optimization}. In our previous work \cite{faraji2014robust}, the inverse dynamics layer could compensate for the inconsistency of LIP trajectories with the real system. In this work, however, thanks to a better consistency of 3LP trajectories, we found inverse kinematics and the physical compliance enough for the range of walking speeds and push recovery scenarios considered. Unperceived locomotion on uneven terrain might require a considerable leg compliance; therefore, we limit the experiments of this work to flat-ground gaits. We only rely on the physical compliance of the robot to absorb impacts and restrict our simple framework to position control. This only requires a simple yet adaptive foot lift strategy to avoid foot scuffing.

\subsection{Novelty}
The novel contribution of this work lies in application of the previously-developed time-projecting controller on the real robot in push recovery and stabilization of walking gaits. We also use the previously-developed 3LP model to generate more natural gaits while keeping the same linear properties of LIP for computations. An extensive set of results demonstrate that the effective control authority is footstep adjustment in our experiments rather than CoP modulation. A considerable portion of the present article is also dedicated to the description of hardware limitations in our robot and the compensatory control strategies which help better realize the 3LP trajectories and the time-projection control. All these control ingredients together motivate application of the proposed method on robots with small feet, soft structural compliance, heavy legs, fast walking frequencies and extreme push recovery conditions.

To implement these ideas on the hardware, we have used well-established methods in the literature for lateral bouncing, leg lift strategies, and torso regulation. The present work is limited to the application of time-projection control on the 3LP gaits. Therefore, we decided to keep a fixed timing instead of a phase-based control, which improves the stability considerably. We also relied on robot's SEA properties while virtual compliance can further smoothen the motion and improve stability. In the following two sections, we describe the 3LP model and the time-projection scheme. Next, we introduce COMAN, discuss sensor qualities and extensively analyze issues arising with SEA elements together with our specific control blocks to compensate them. Open-loop and closed-loop control details are also described in the same section. Different scenarios of in-place walking, intermittent push recovery, continuous push recovery, walking gait and ankle stiffness measurements are presented in the results section. Finally, the paper is concluded by a discussion of results and possible future improvements.

\section{3LP Model}

\begin{figure}[]
	\centering
	\includegraphics[trim = 0mm 0mm 0mm 0mm, clip, width=0.3\textwidth]{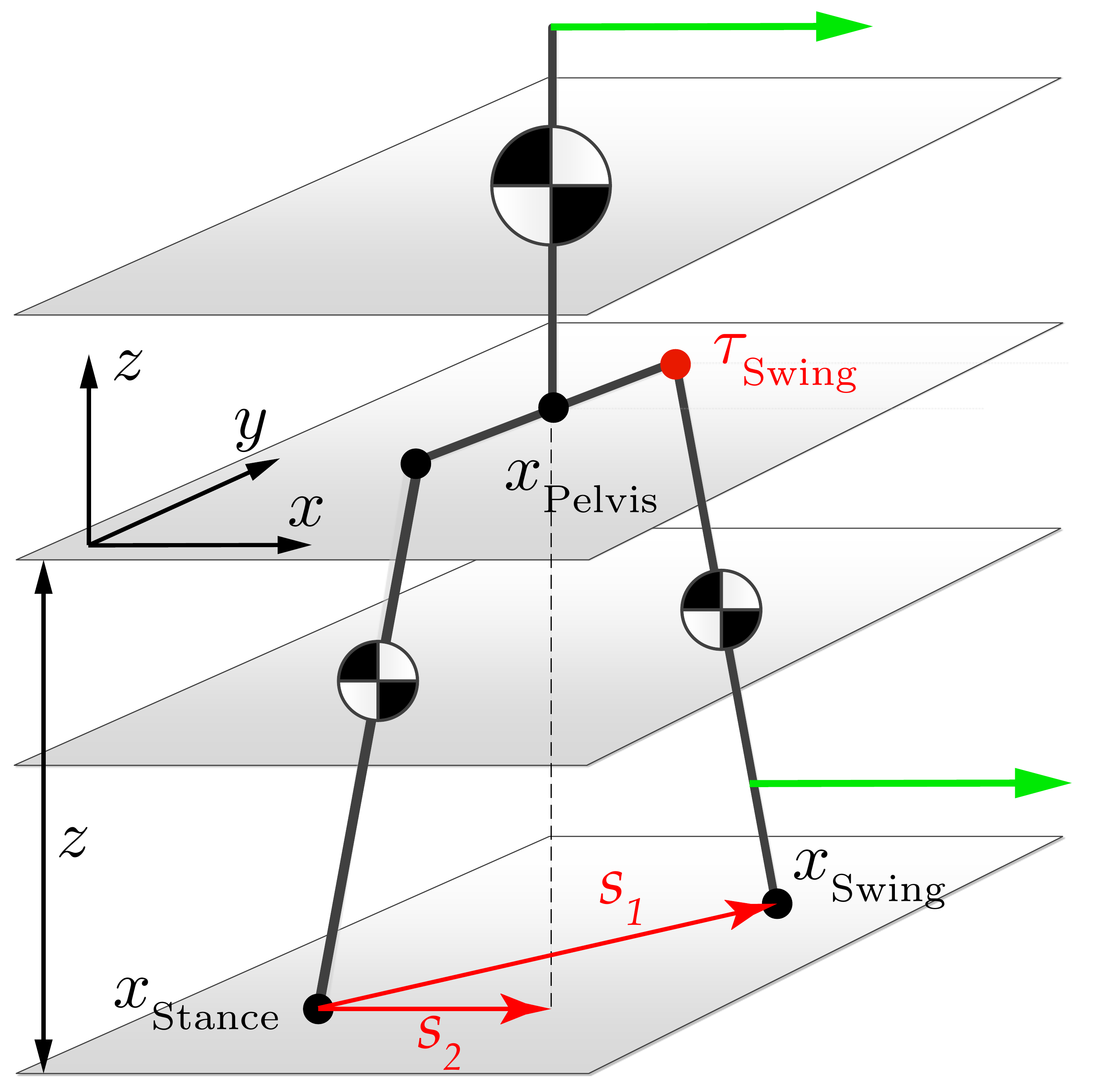}
	\caption{A schematic of 3LP model used in this paper for walking gait generation \cite{faraji20173lp}. This 3D model is composed of three linear pendulums connected with a massless pelvis of a certain width to simulate natural lateral bounces. All masses and the pelvis stay in constant-height planes which make the model linear. The torso also remains upright by an ideal actuator placed in the stance hip joint. The state of this model is described by pelvis and swing foot horizontal positions while the swing hip and stance ankle torques serve as inputs to generate different walking gaits.} 
	\label{fig::3lp}
\end{figure}

The 3LP model, as its name implies \cite{faraji20173lp}, is composed of three linear pendulums to simulate legs and torso dynamics in walking. These pendulums are connected through a mass-less rigid pelvis of a certain width which stays in a constant-height plane, similar to the LIP model. Each pendulum approximates the entire corresponding limb in the robot with a mass in the middle and inertia in the sagittal and lateral planes, but not around the pendulum rod itself. All masses stay in certain constant-height planes, assuming ideal prismatic actuators in the legs like the LIP model. Apart from the prismatic actuators, 3LP has three groups of two rotary actuators for lateral and sagittal directions, placed in the stance hip, swing hip and stance ankle joints. This model simulates no pelvic and torso rotation, assuming ideal stance hip actuators to compensate the internal coupling between the limbs. Figure \ref{fig::3lp} demonstrates the 3LP model conceptually with the three pendulums, limb masses, fixed-height planes, state variables, actuation dimensions and relative position vectors $s_{1}$ and $s_{2}$ (described later).

\subsection{Mechanics}

In this article, we skip complicated mechanical equations and refer to the original paper \cite{faraji20173lp} where the 3LP model was introduced. In brief, Newtonian equations of motion are written for each limb (and the pelvis) with incoming and outgoing forces and torques. Swing hip and stance ankle actuators are used as inputs. We consider the torques in these actuators as free variables while all other internal forces and torques are resolved by a symbolic combination of mechanical equations. In this paper, we only use hip torque modulation though and leave the ankle joints free, i.e., producing a compass gait. The model state in 3LP is composed of horizontal pelvis and feet positions which together form a vector $x(t) \in \mathbb{R}^6$. The input vector $u(t) \in \mathbb{R}^2$ also represents swing hip torques, defined as:
\begin{eqnarray}
	x(t) = \begin{bmatrix} x_{Swing}(t) \\ x_{Pelvis}(t) \\ x_{Stance}(t) \end{bmatrix},\ \ u(t) = \tau_{Swing}(t)
\end{eqnarray}
where variables are depicted in Figure \ref{fig::3lp}. Equations of motion are found as:
\begin{eqnarray}
	\frac{d^2}{dt^2}{x}(t) = C_x \ x(t) + C_u \ u(t) + C_d \ d
	\label{eqn::continuous_3lp}
\end{eqnarray}
where $d=\pm1$ determines left or right support phase. The inclusion of the variable $d$ is only because of the non-zero pelvis width which distinguishes between left and right bouncing. Note that the stance foot is assumed to be stationary. Therefore, the matrix $C_x$ does not influence $x_{Stance}(t)$ and its derivative (which is equal to zero), ensuring $\ddot{x}_{Stance}(t) = 0$ during the swing phase. After each phase of motion, the swing and stance feet can be exchanged through multiplying $x(t)$ by the following matrix:
\begin{eqnarray}
	S_x =& \begin{bmatrix} 	\cdot & \cdot & I_{2\times2} \\
	\cdot & I_{2\times2} & \cdot \\
	I_{2\times2} & \cdot & \cdot 
	\end{bmatrix}
\end{eqnarray}

In the original 3LP paper \cite{faraji20173lp}, we also derived equations for a double support phase by considering linear contact force transitions. In the present paper, however, we remove this phase for simplicity. We will later show that the intrinsic physical compliance of COMAN naturally produces a double support phase of approximately $20\%$ in each phase.

\subsection{Closed-Form Equations}

By construction, the 3LP equations are linear with respect to the state variables and inputs. To obtain closed-form equations that describe the system evolution in future, it is enough to consider certain swing hip torque profiles and solve the second order differential equations of (\ref{eqn::continuous_3lp}). We consider simple piecewise liner profiles which give us enough freedom for control and closely approximate human profiles \cite{faraji20173lp}:
\begin{eqnarray}
u(t) = u_{c} + t\ u_{r}
\end{eqnarray}
where $u_{c}, u_{r} \in \mathbb{R}^2$ are constant and time-increasing terms in both sagittal and lateral directions. With these definitions, the closed-form 3LP equations are obtained as:
\begin{eqnarray}
q(t) = A(t) q(0) + B(t) u + C(t) d
\label{eqn::discrete_3lp}
\end{eqnarray}
where $q(t) \in \mathbb{R}^{12}$ and $u \in \mathbb{R}^4$ are:
\begin{eqnarray}
q(t) = \begin{bmatrix} x(t) \\ \dot{x}(t) \end{bmatrix} ,\ u = \begin{bmatrix} u_{c} \\ u_{r} \end{bmatrix}
\end{eqnarray}

\subsection{Periodic Gaits}

Knowing the state evolution equations of 3LP in (\ref{eqn::discrete_3lp}), we can find successive phases of motion which are symmetric with respect to each other. Consider the matrix:
\begin{eqnarray}
	M_x =& \begin{bmatrix}   I_{2\times2} & \cdot & -I_{2\times2} \\
	\cdot & I_{2\times2} &-I_{2\times2} 
	\end{bmatrix} 
	\label{eqn::symmetry}
\end{eqnarray}
which extracts relative position vectors $s_{1}$ and $s_{2}$ from the vector $x(t)$, depicted in Figure \ref{fig::3lp}. At the beginning of each phase, these relative positions are the same in the sagittal direction and opposite in the lateral direction. Besides, the initial and final swing foot velocities together with the stance foot velocity should be zero (we assume impact-less locomotion in 3LP). Consider matrices $S$, $M$, $N$ and $O$ as:
\begin{eqnarray}
	\nonumber S =& \begin{bmatrix} S_x & \cdot \\ \cdot & S_{x} \end{bmatrix}, \quad 	M = \begin{bmatrix} M_x & \cdot \\ \cdot & M_{x} \end{bmatrix} \\
	\nonumber N =& \begin{bmatrix}   \cdot & \cdot & \cdot & I_{2\times2} & \cdot & \cdot  \\ \cdot & \cdot & \cdot & \cdot & \cdot & I_{2\times2}  \end{bmatrix} \\
	O =& diag(\ [1,-1, 1,-1, 1,-1, 1, -1]\ )
\end{eqnarray}
where $N$ implements initial foot velocity conditions and $O$ imposes gait symmetry. A valid periodic initial condition $\bar{Q}[k]$, inputs $\bar{U}[k]$ and direction $\bar{D}[k]$ must satisfy the following equations:
\begin{eqnarray}
	\nonumber M \bar{Q}[k] &=& OMS(A(T)\bar{Q}[k]+B(T)\bar{U}[k]+C(T)\bar{D}) \\
	N \bar{Q}[k] &=& 0
	\label{eqn::3lp_periodic}
\end{eqnarray}
which evolve $\bar{Q}[k]$ for one phase according to (\ref{eqn::discrete_3lp}), exchange the feet by $S$, extract relative vectors $s_{1}$ and $s_{2}$ and their derivatives by $M$ and apply the symmetry concept by $O$. The underlying assumption is a fixed gait frequency determined by step time $T$. In this paper, we use capital letters to show variables that are treated in discrete equations. The null-space formed by equations (\ref{eqn::3lp_periodic}) still leaves four degrees of freedom. The desired average walking velocity takes two of these dimensions while we resolve the other two by a minimization of hip torques. In other words, among all possible combinations of $\bar{Q}[k]$ and $\bar{U}[k]$ (for a fixed $\bar{D}[k]=1$ for example) which form a space of four dimensions, we find a combination with the desired average speed and the minimal hip torque magnitudes (refer to \cite{faraji20173lp} for further details). The open-loop 3LP gait kinematics $\bar{q}(t)$ used in our controller is therefore defined as:
\begin{eqnarray}
	\bar{q}(t) = A(t) \bar{Q}[k] + B(t) \bar{U}[k] + C(t) \bar{D}[k]
	\label{eqn::nominal_solution}
\end{eqnarray}
at each instance of time $t$ and walking phase $\bar{D}[k]=\pm 1$.

\section{Time-Projection Control}

3LP model provides closed-form equations (\ref{eqn::discrete_3lp}) that describe system evolution over successive motion phases. Thanks to linearity, it is very simple to find a Poincar\'e map or linearized discrete model around the walking gait we obtained in the previous section. In this regard, we consider the same phase-change event, add an error to the state vector, evolve the erroneous state and a delta input until the end of the phase and measure the new error. Consider the stack of relative position vectors $s_1(t)$ and $s_2(t)$ and their derivatives which are extracted from the full vector by $s(t)=Mq(t)$ where $s(t) \in \mathbb{R}^8$. Consider also a matrix $\hat{M}$ defined as:
\begin{eqnarray}
	\hat{M} =& \begin{bmatrix} \hat{M}_x & \cdot \\ \cdot & \hat{M}_x	\end{bmatrix}, \quad
	\hat{M}_x =& \begin{bmatrix}   I_{2\times2} & \cdot \\
								   \cdot & I_{2\times2} \\ 
								   \cdot & \cdot \\ 
	\end{bmatrix} 
	\label{eqn::symmetry_reverse}
\end{eqnarray}
which distributes an error $e(t)$ on $s(t)$ to the full state vector $q(t)$. Since the relative vectors are extracted from the full vector by matrix $M$, the product $M \hat{M}$ naturally equals to identity. An initial error vector $E[k]$ added to $\bar{Q}[k]$ by $Q[k] = \hat{Q}[k]+\hat{M}E[k]$ evolves in time according to (\ref{eqn::discrete_3lp}) where an additional input $\Delta U[k]$ is also given to the system. The evolved error can be extracted from the next discrete state by:
\begin{eqnarray}
	E[k+1] = OMS(Q[k+1] - \bar{Q}[k+1])
	\label{eqn::error3}
\end{eqnarray} 
which leads to the following linear discrete model:
\begin{eqnarray}
\nonumber E[k+1] =& \hat{A}(T) E[k] + \hat{B}(T) \Delta U[k] \\
\hat{C} E[k+1] =& 0
\label{eqn::error_dynamics}
\end{eqnarray} 
where:
\begin{eqnarray}
\nonumber \hat{A}(T) =& O M S A(T) \hat{M} \\
\nonumber \hat{B}(T) =& O M S B(T) \\
\hat{C} =& \begin{bmatrix} 0_{2\times4} & I_{2\times2} & 0_{2\times2} \end{bmatrix}
\end{eqnarray}
and the matrix $\hat{C}$ is defined to constrain final swing foot velocities to zero.

\subsection{Discrete LQR Control}

Knowing error evolution equations (\ref{eqn::error_dynamics}) and the effect of inputs, we can design a Discrete LQR controller (DLQR) to stabilize the system. Our particular system must satisfy a constraint on the foot velocity too which requires a simple manipulation of equations and DLQR's cost function. We refer to \cite{faraji20173lp2} for further details and only take the resulting control gain matrix $K$ which produces a corrective feedback $\Delta U[k] = -K E[k]$. A DLQR controller can be triggered at the beginning of each phase and produce a delta actuator input to correct the error at the end of the phase. Due to a delayed reaction, however, this controller cannot reject intermittent disturbances optimally \cite{faraji20173lp2}. Therefore, we use a time-projecting controller that can react to inter-sample disturbances immediately by using the expertise of the DLQR controller. This idea is briefly introduced in the next section.

\subsection{Continuous Time-Projection}

\begin{figure*}[]
	\centering
	\includegraphics[trim = 0mm 0mm 0mm 0mm, clip, width=1\textwidth]{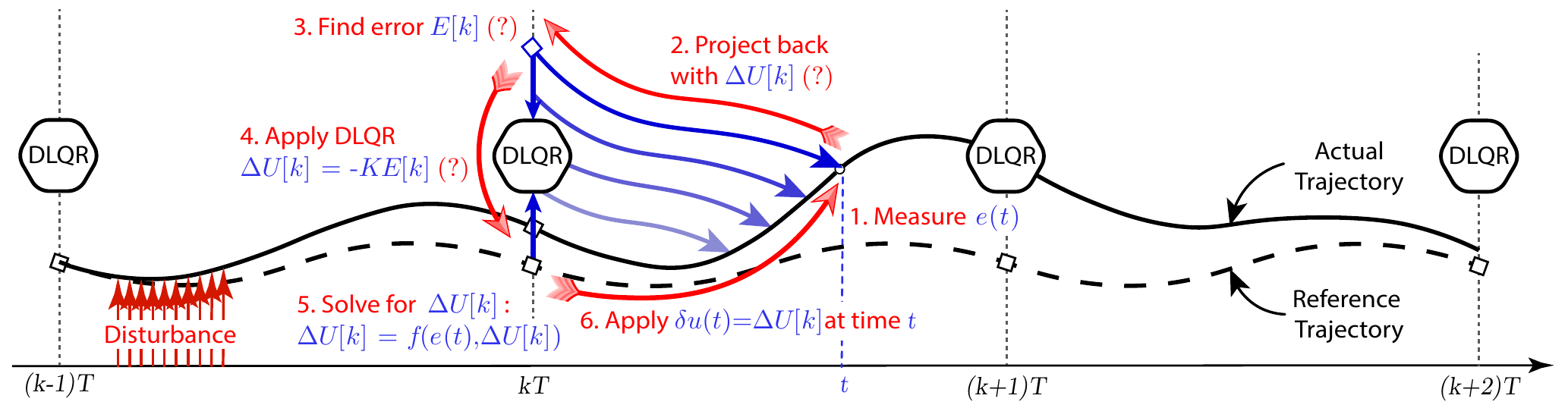}
	\caption{Time-projection control over a nominal system trajectory. Imagine a control period $T$ is decided for which, a DLQR controller is calculated to stabilize the system. For the walking application, we set $T$ equal to the actual step time. The DLQR controller can only be triggered at time instances $kT$ which makes it sensitive to inter-sample disturbances, especially if the system has largely unstable modes. However, we can take advantage of DLQR knowledge and provide immediate corrections at every inter-sample time $t$. Knowing system evolution matrices from the closest discrete sample $kT$ until time $t$, we can project the measured error $e(t)$ back in time and obtain an equivalent discrete error $E[k]$. This error can evolve in time with some $\Delta U[k]$ and lead to the currently observed error. To resolve the ambiguity between $E[k]$ and $\Delta U[k]$, we link them via DLQR together in step 4 of time-projection. Therefore, DLQR provides a stabilizing input which we directly apply to the system at time $t$.} 
	\label{fig::projecting}
\end{figure*}

To formulate the time-projection idea, we consider a free system without constraints to present simpler formulations and provide an easier understanding. Handling time-projection for constrained systems can be found in the original control paper \cite{faraji20173lp2}. Having solved 3LP equations in a closed-form and derived discrete error dynamics, we showed that we can easily find a DLQR controller that stabilizes the system by swing hip torque adjustments and in consequence, footstep adjustments. Now consider an on-line control paradigm in which we can measure the system error at any inter-sample time and react to it quickly. The reaction might not stabilize the system immediately like CoP modulation though, but a proper adjustment of swing foot location can considerably save control effort in the following phases. This is because the current weakly coupled swing position later becomes the new stance foot position which tightly couples to other system variables. 

Consider an inter-sample time $0\le t \le T$, the nominal periodic solution $\bar{q}(t)$ defined in (\ref{eqn::nominal_solution}), the current measured state $q(t)$ and the instantaneous error:
\begin{eqnarray}
e(t) = M(q(t)-\bar{q}(t))
\label{eqn::current_error}
\end{eqnarray}
We can apply a corrective swing hip torque $\delta u(t)$ at time $t$, in addition to the nominal actuator inputs $\bar{u}(t) = \bar{u}_c + t \bar{u}_r$. This scenario is depicted in Figure \ref{fig::projecting} in details. The time-projection controller takes the following steps in sequence to find the vector $\delta u(t)$:
\begin{enumerate}
	\item Measure the current numeric error $e(t)$ at time $t$. 
	\item Consider an unknown parametric input $\Delta U[k]$ until time $t$ and project the measured error $e(t)$ back in time.
	\item Calculate a possible parametric initial vector $E[k]$ by:
	\begin{eqnarray}
	e(t) = A(t-kT) E[k] + B(t-kT) \Delta U[k]
	\end{eqnarray}
	\item Apply DLQR on $E[k]$ to find $\Delta U[k]$:
	\begin{eqnarray}
	\Delta U[k] = -K E[k]
	\end{eqnarray}
	\item Now find $\Delta U[k]$ by solving a linear system of equations:
	\begin{eqnarray}
	\begin{bmatrix} A(t-kT) & B(t-kT) \\ K & I \end{bmatrix} \begin{bmatrix}
	E[k] \\ \Delta U[k] \end{bmatrix} = \begin{bmatrix} e(t) \\ \cdot
	\end{bmatrix}
	\label{eqn::simple_system_solvedU}
	\end{eqnarray}
	\item Assuming $\delta u(t) = \Delta U[k]$, take the resulting corrective input and apply it to the system at time $t$.
\end{enumerate}
The time-projection control involves solving a linear system of equations at every control sample which only takes few microseconds on a modern computer, compared to at least few hundred microseconds taken by our previous MPC controller introduced in \cite{faraji2014robust}. Note that the projecting controller does not need to know disturbing forces while it produces hip torques which are integrated to produce footstep adjustments. In the absence of errors $e(t)$, this controller produces no correction. In these conditions, it produces a constant $\delta u(t)$ which is equal to a discrete correction that the DLQR controller would produce at the beginning of the phase. In other words, as long as the system evolves without inter-sample disturbances, the projecting controller has no advantage over the DLQR controller in providing on-line corrections. In our previous work \cite{faraji20173lp2}, we extensively discuss the recovery performance against different inter-sample push timings as well as analysis of controllable regions, given certain hardware limitations. Due to an online numerical optimization in MPC, it can handle such inequality constraints easily. The time-projecting controller is blind to these constraints. However, our analysis shows that in normal walking conditions, time-projection controller covers most of the whole set of controllable states. Therefore, MPC might only cover a slightly bigger set of states that are not visited most of the time in practice \cite{faraji20173lp2}.

\section{Hardware Control}

In this section, we explain implementation details of our controller on the real hardware, describe limitations and motivate individual control blocks used to track 3LP gaits and stabilize them via time-projection. As mentioned earlier, COMAN has relatively short and heavy legs which motivate considering swing dynamics at the gait generation level. However, SEA actuators and hardware compliance improve stability and performance in perturbed walking conditions. We believe performing torque control might bring additional compliance, which is beneficial, however low signal qualities and the hardware's mechanical status to date do not allow a convincing tracking performance. Therefore, we limited our focus to a position control paradigm.

\begin{figure*}[]
	\centering
	\includegraphics[trim = 0mm 0mm 0mm 0mm, clip, width=1\textwidth]{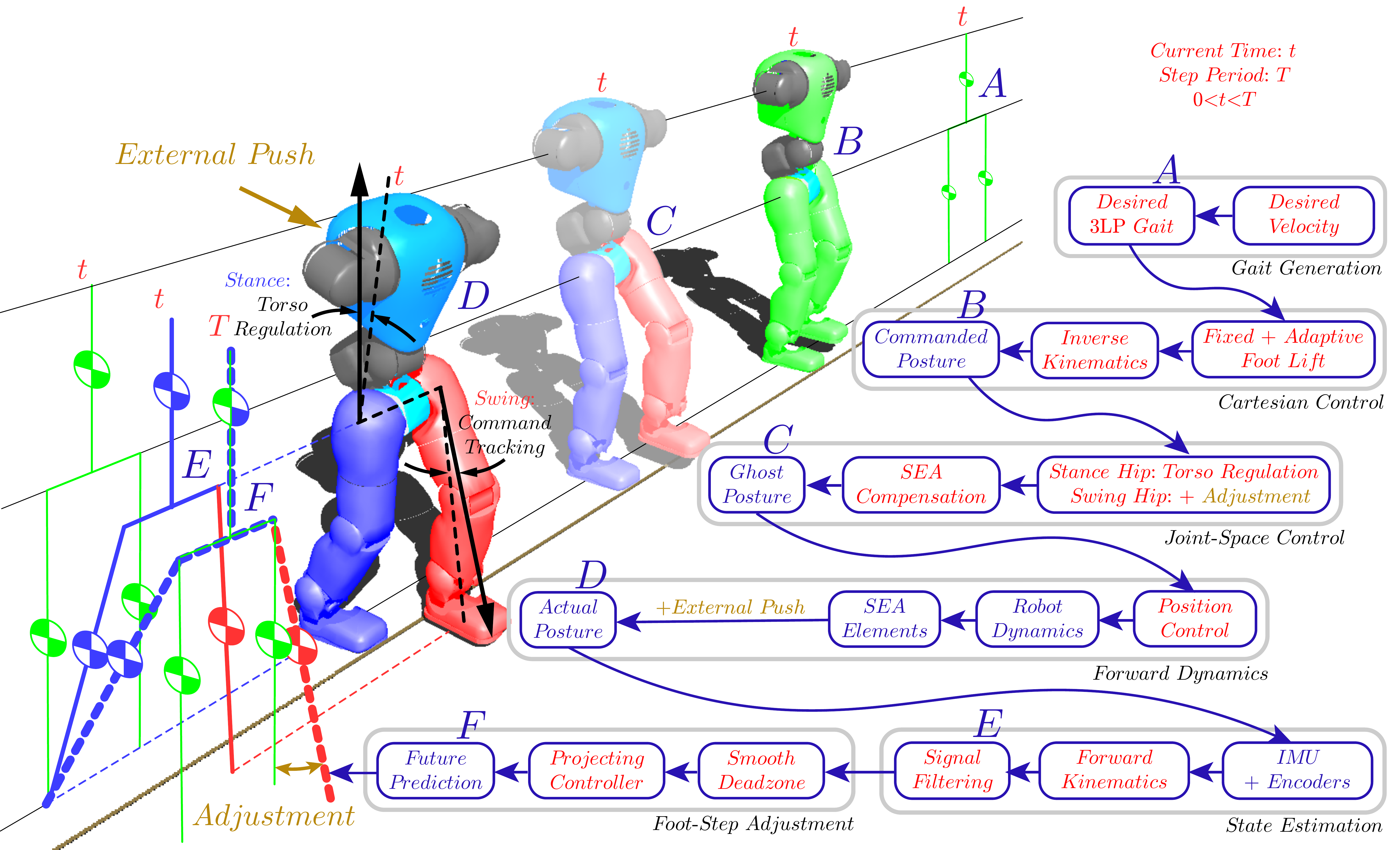}
	\caption{The entire control architecture proposed in this paper. Given a desired walking speed, 3LP produces a nominal gait $A$ which describes the motion of pelvis and both feet in the Cartesian space. We take these trajectories, add fixed and adaptive foot lift components to the swing foot and pass them through an inverse kinematics layer which finds desired joint angles $B$. The swing hip follows the desired joint angle while the stance hip follows the desired torso angle which is realized by a smooth switch between swing hip encoder and IMU on the pelvis. A time-projection based adjustment is also added to the swing hip trajectory. We add a fixed small angle to the stance knee trajectory to compensate compression of SEA springs in the knee. Joint trajectories are tracked by relatively stiff position controllers on motor shafts before the springs. This creates a posture $C$ which is called the ghost robot, referring to pre-spring motor positions. The actual link positions $D$ are defined by post-spring encoders though, passed through a forward kinematics layer to obtain actual Cartesian feet and pelvis positions. The IMU signal is also used at this stage to orient the robot properly. After a relatively strong filtering of Cartesian positions to remove noise, we apply a dead-zone function too which makes them clean for time-projection control. We take the currently filtered state $E$, simulate it until end of the phase by a closed-loop time-projection controller to obtain the posture $F$. We measure the final attack angle adjustment (with respect to the nominal gait) and add it to the swing hip trajectory then. If an external push is applied, the instantaneous error (after filtering) produces a proper attack-angle adjustment through time-projection which captures the push in few steps.} 
	\label{fig::control}
\end{figure*}

Figure \ref{fig::control} shows the big picture of our entire control framework with all individual components. The 3LP Cartesian gaits are augmented by foot lift strategies and converted to joint angles through inverse kinematics. Hip switching and SEA compensation blocks further augment joint-space trajectories and make them suitable for position control. Actual link encoder and IMU signals are then used for a state estimation whose output goes to the time-projecting controller. This controller simulates the instantaneous measured state until the end of the phase and takes the resulting footstep adjustment to augment the swing hip angle found in earlier control stages. Such adjustment stabilizes the motion, especially in the presence of external disturbances. In the rest of this section, we discuss each control block in details. 

\subsection{Low-Level Challenges}

COMAN (height of $\approx90\ cm$ and mass of $\approx30\ kg$) has $23$ actuated joints and is equipped with pre-spring and post-spring encoders, strain gauge torque sensors, a high-end IMU and two 6D contact force sensors \cite{coman2}. It offers a proper level of compliance with the SEA elements which are soft enough to absorb impacts and stiff enough to keep the posture upright. However, the quality of sensory data, actuation limitations, joint backlashes and extra spring deflections make the control of walking at a relatively high-frequency choice of $2.5\ steps/s$ more challenging. 

\subsubsection{Sensor Quality:}

\begin{figure*}[]
	\centering
	\includegraphics[trim = 0mm 0mm 0mm 0mm, clip, width=1\textwidth]{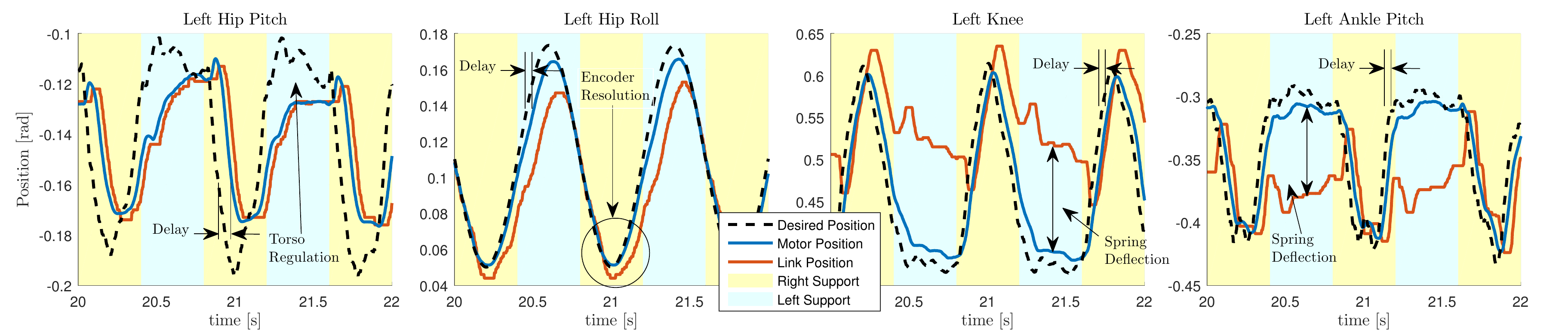}
	\caption{Typical trajectories of joints involved in an in-place walking gait. Link encoders produce noisy signals which cannot be used for stiff position control directly, but motor encoders are precise enough for this purpose. In the hip pitch joints, we use a switching control rule in swing and stance phases which applies footstep adjustment and corrects for IMU angles respectively. Due to the noise on these signals, we use a smaller position gain in hip pitch joints which increases the tracking delay. Considerable spring compressions are also observed in the knee and ankle pitch joints which increase the mismatch between actual and ghost robots in practice. } 
	\label{fig::dealys_resolution}
\end{figure*}

COMAN has incremental optical encoders to measure pre-spring motor shaft positions and absolute differential magnetic encoders to measure post-spring link positions. The optical encoder signal has a good quality due to a division by harmonic drive's reduction ratio ($N=101$). The absolute encoder, however, covers the whole span of $360^o$ by only $12$ bits which produce a considerable quantization error. Dedicated actuator control boards are connected to local LAN dispatchers which are all connected in a tree-shaped topology whose root goes to the external controlling PC. This network architecture together with internal low-level motor controllers lead to a delayed position tracking of at least $20\ ms$ in the stiffest and aggressive conditions. Due to the quality of link encoders, delays, backlashes and the fact that we do not want to "fight" against the physical compliance to benefit from impact absorption properties, we perform position control only on the motor shaft (and not the link position) with proper signals from the optical encoder. Figure \ref{fig::dealys_resolution} shows sensor qualities and delays in an in-place walking task and important joints involved.

\subsubsection{The Ghost Robot:}

\begin{figure}[]
	\centering
	\includegraphics[trim = 0mm 0mm 0mm 0mm, clip, width=0.35\textwidth]{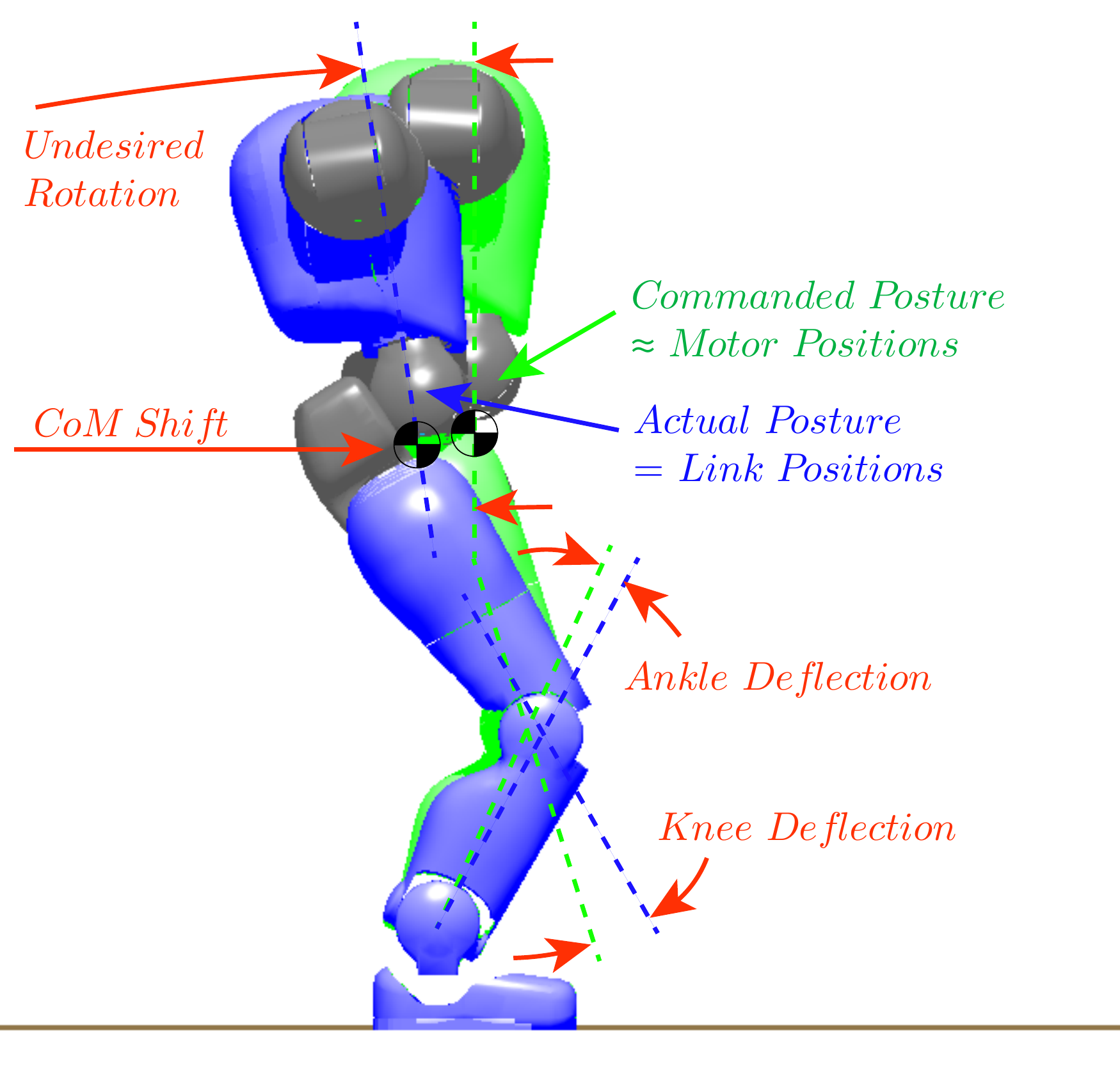}
	\caption{When commanding a resting posture with the CoP in the middle of support polygon, spring compressions lead to a different posture. Ankle spring compressions make the robot lean forward while knee spring compressions make it lean backward. These two effects cancel each other to some extent, but in practice, the torso would have an undesired rotation depending on desired knee angles and CoP locations. The actual CoM might also move backward for $1-2\ cm$.} 
	\label{fig::ghost}
\end{figure}

As mentioned earlier, SEA compliance in COMAN is enough to let it keep balance to some extent, but depending on the commanded knee angles, spring compressions produce an error in tracking. Imagine we command a crouched standing posture shown in Figure \ref{fig::ghost} with the CoP in the middle of the feet. Due to the ankle pitch compliance, the robot tends to lean forward while due to the knee compliance, the robot leans backward. These two effects slightly cancel each other, but in practice, the knee deflection systematically perturbs the state estimation more. This is because the relative horizontal feet-pelvis position highly depends on the knee angle (with a radius equal to the shank length), but much less on the ankle angle. Thus, the mismatch between actual and ghost postures is small but needs to be considered in control design. Note that we solely rely on the pelvis IMU for state estimation. An alternative would be to assume a flat foot on the ground and to build the kinematic chain upwards. We avoid this approach to minimize the effect of ankle spring deflections, slippages and foot tilt/roll effects. Because of a particular actuator design, a backlash appears in the joints of COMAN over time which requires regular maintenance. At the time of experiments, we had significant backlashes in the knee and ankle joints which were partially removed by external rubber bands (as shown in the Multimedia Extension 1). However, these inaccuracies further complicated the state estimation described later in this section.

\subsubsection{Actuator Limitations:}

\begin{figure}[]
	\centering
	\includegraphics[trim = 5mm 0mm 10mm 0mm, clip, width=0.5\textwidth]{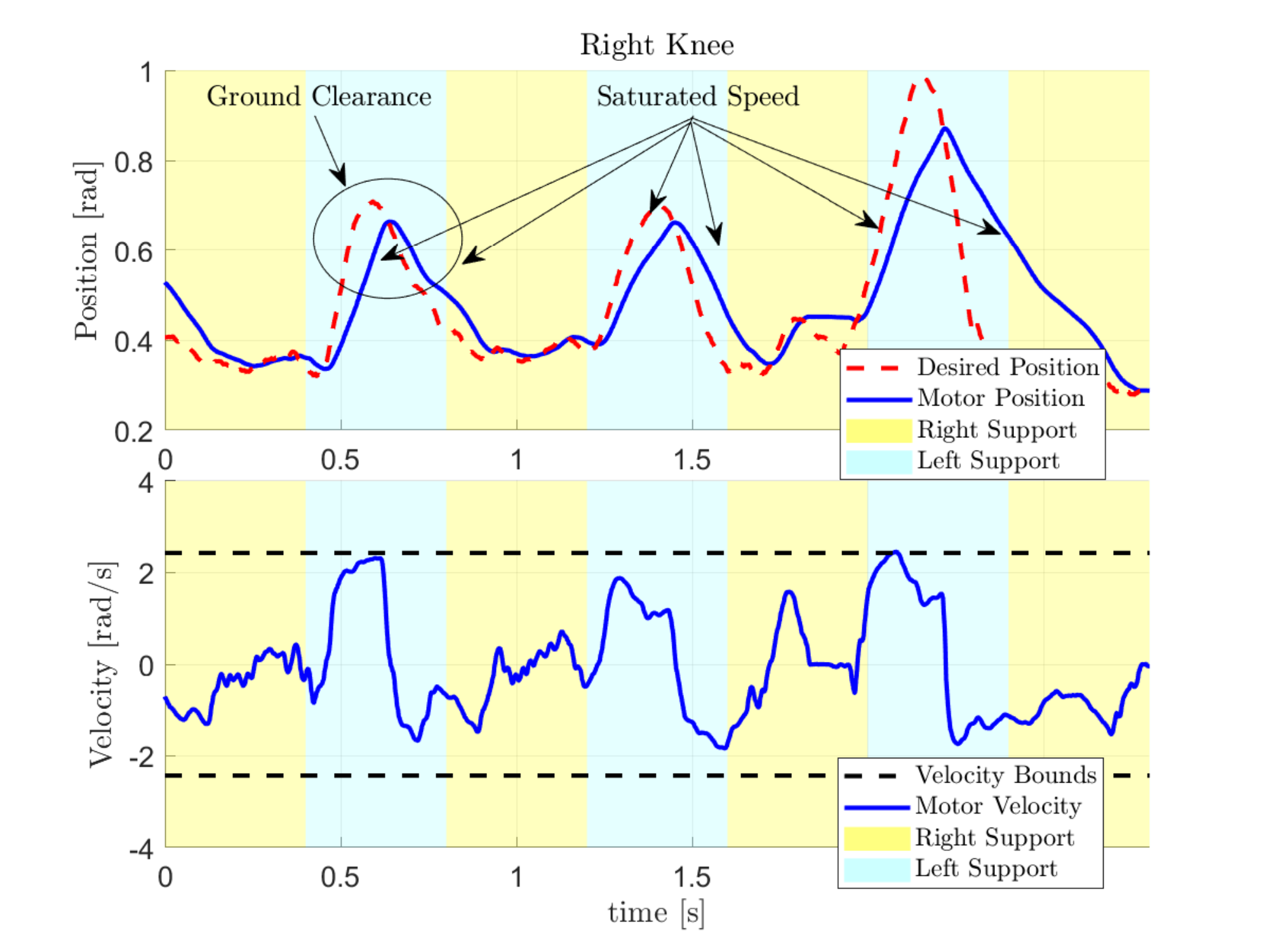}
	\caption{In normal or perturbed walking conditions, a small leg lift of $2-3\ cm$ cannot be easily achieved due to a saturation of actuator velocities. The maximum velocity is determined by power supply voltage and motor properties. Due to the targeted perturbed walking conditions in our experiments, we operate the robot with a slightly lower but safe voltage to avoid high currents and unwanted shut-downs. This limits the leg lift performance which might lead to falling when a fast swing progression is commanded, but the foot is not lifted enough, and the toes still touch the ground. } 
	\label{fig::max_speed}
\end{figure}

In a previous work \cite{faraji2015practical}, to perform whole-body inverse dynamics and torque control, we attempted to identify a model for COMAN actuators. We designed training trajectories with static and dynamic profiles and used a simple proportional controller to track these trajectories by producing actuator voltages. By measuring the output torque, we were able to relate actuator voltages, velocities and output torques together:
\begin{eqnarray}
\nonumber v(t) =& Ri(t) + kN\dot{\theta}(t) \\
kNi(t) =& \tau_{out}(t) + A_v \dot{\theta}(t) + A_c |\dot{\theta}(t)| + J_m \ddot{\theta}(t)
\label{eqn::motor_vi}
\end{eqnarray}
where $v(t)$ denotes motor voltage, $R$ denotes terminal resistance, $i(t)$ denotes motor current, $k$ denotes torque constant which is assumed to be equal to the back-emf constant, $N$ is harmonic drive ratio, $\tau_{out}(t)$ is output torque, $\theta(t)$ is shaft position, $J_m$ is reflected rotor inertia and $A_v$ and $A_c$ are coefficients of a simple Coulomb-Viscous friction model. Since current measurements were not reliable enough, we combined equations of (\ref{eqn::motor_vi}) and obtained:
\begin{eqnarray}
v(t) = \alpha \dot{\theta}(t) + \beta  (J_m \ddot{\theta}(t) + \tau_{out}(t) + A_c |\dot{\theta}(t)|)
\end{eqnarray}
where:
\begin{eqnarray}
\nonumber \alpha = \frac{RA_v}{kN}+kN, \quad \beta = \frac{R}{kN}
\end{eqnarray}
In COMAN actuators on average $\alpha \approx 5.42\ Vs/rad$, $\beta \approx 0.45\ V/Nm$, $J_m \approx 0.23\ Nms^2/rad$ and $A_c \approx 1.66\  Nms/rad$. This means voltages are converted to joint torques with a coefficient of $\beta \approx 0.45\ V/Nm$ and there is a damping of $\alpha / \beta \approx 12\ Nms/rad$ in practice. A challenge in our experiments was to operate the robot in a reduced maximum voltage of $15\ V$ which leads to a theoretic velocity bound of $\dot{\theta}_{max}=\pm2.43\ rad/s$. Regarding the choice of walking frequency ($2.5\ steps/s$), the knees do not have enough time to bend fast which makes the knee-bending foot lift strategy limited. We have demonstrated this effect over some perturbed gait trajectories in Figure \ref{fig::max_speed} where fixed and adaptive leg lift strategies (described later) command a fast bending angle to the right knee. We normally operate the robot with knee angles of $0.4\ rad \approx 23^o$ while, as depicted in Figure \ref{fig::max_speed}, a leg lift of $2-3\ cm$ cannot be completely realized due to saturated knee velocities. In practice, we cannot operate the robot in straight-knee postures either, although our inverse kinematics is able to handle singularities \cite{faraji2017singularity}. Limited velocities motivated us to combine a hip-roll foot lift strategy with the knee-bending strategy to provide an extra leg left without overloading the knee joints. A delayed leg lift is hazardous in practice, for example when the hip is pushing for a fast swing trajectory, but the foot is yet not lifted enough to clear the ground. In this case, the toes might touch the ground much earlier than expected which may lead to an immediate fall.

\subsubsection{SEA Compensation:}

\begin{figure}[]
	\centering
	\includegraphics[trim = 0mm 0mm 0mm 0mm, clip, width=0.5\textwidth]{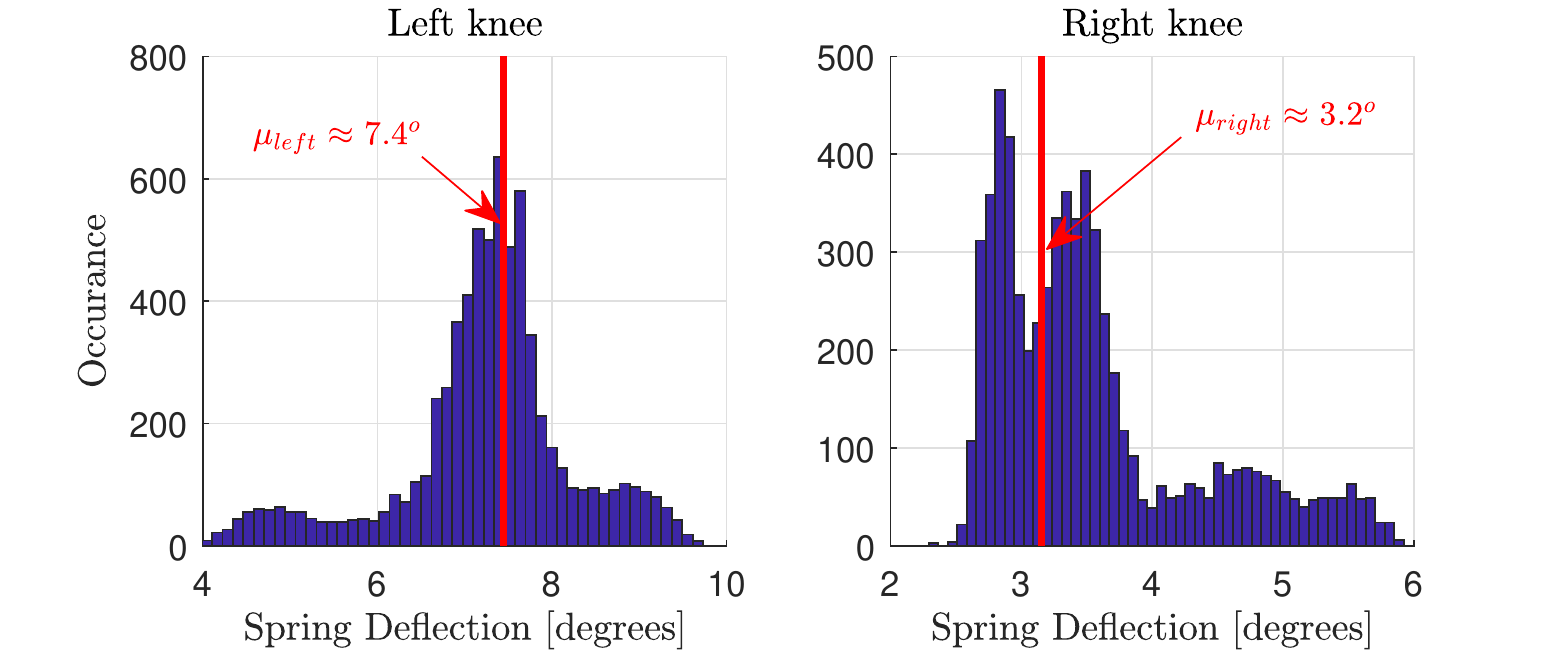}
	\caption{Histograms of typical knee spring compressions during an in-place walking gait when experiments where done. These plots depend on the status of springs because they permanently deflect over time and lose their original stiffness which needs a regular replacement. Assuming short step lengths, we compensate these deflections by a feed-forward angle added to the desired knee trajectories. The compensating angles $\mu_{left}$ and $\mu_{right}$ are constant all the time, but multiplied by the contact forces to make them effective only during the stance phase in each leg.} 
	\label{fig::knee_deflection}
\end{figure}

As described earlier, due to the spring compressions, the ghost robot might mismatch the actual robot considerably. This mismatch is hard to compensate in the ankles since the CoP moves back and forth easily in practice and the compression angle is variable. In the knees, however, this compression is quite constant assuming relatively short step lengths. Due to heavy duty daily operations, the knee springs get permanently deflected in COMAN over time which requires frequent replacement. In this regard, we perform an open-loop in-place walking gait before experiments to identify steady-state knee spring compressions which are plotted as histograms in Figure \ref{fig::knee_deflection}. This test can be easily repeated to identify the current state of the robot before experiments. In our algorithm, the statistical average of these histograms is used as a constant, multiplied by normalized contact forces and added to the desired knee angles. In other words, assuming $F_{left}(t)$ and $F_{right}(t)$ to represent instantaneous filtered contact force measurements, $mg$ to be the total weight and $\mu_{left}$ and $\mu_{right}$ to represent mean deflections, we add $\delta \theta_{left}(t)$ and $\delta \theta_{right}(t)$ to the desired knee trajectories: 
\begin{eqnarray}
\nonumber \delta \theta_{left}(t) = -\frac{|F_{left}(t)|}{mg} \mu_{left} \\
\delta \theta_{right}(t) = -\frac{|F_{right}(t)|}{mg} \mu_{right}
\label{eqn::sea_compensaiton}
\end{eqnarray}

These adjustments can be a function of time as well, but we found this adaptive compensation smoother. Besides, the magnitude of adjustment is so small that it does not cause shaking when directly combined with contact forces and given to the stiff position controllers. However, the compensatory influence is considerable. Given a shank length of $\approx 25\ cm$, $\delta \theta_{left}(t)$ and $\delta \theta_{right}(t)$ each compensate about $3.2\ cm$ and $1.4\ cm$ of errors on relative horizontal pelvis-feet positions respectively.

\subsection{Open-Loop Control}

This section introduces simple strategies used to generate a periodic rhythm of motion based on 3LP gaits. The components of our open-loop control are gait generation by 3LP, foot lift strategies and hip switching rules.

\subsubsection{Lateral Plane:}

\begin{figure}[]
	\centering
	\includegraphics[trim = 0mm 0mm 0mm 0mm, clip, width=0.35\textwidth]{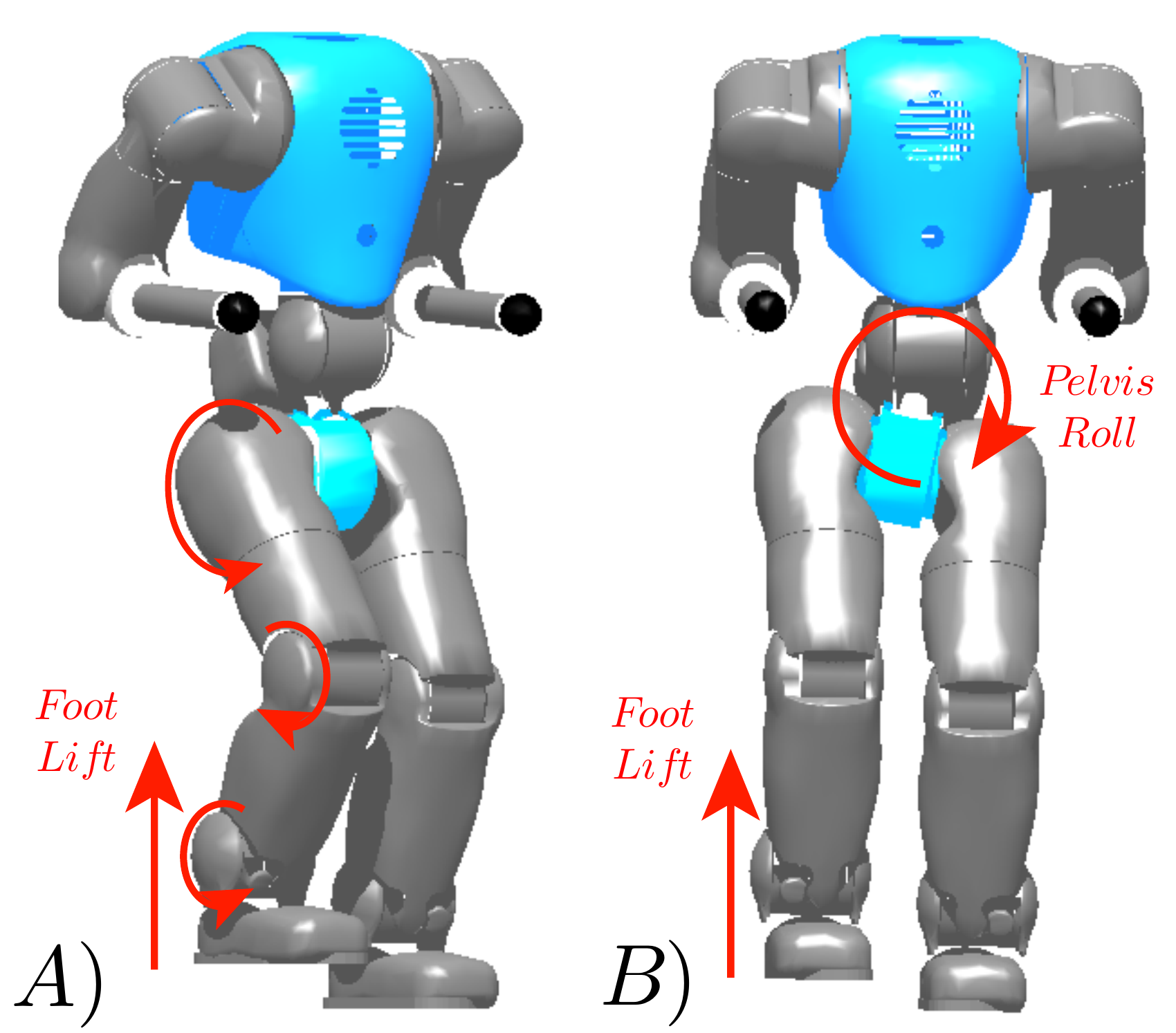}
	\caption{A) Knee-bending and B) hip-roll strategies used to lift the foot in our walking gait generation. In practice, an open-loop foot lift of about $2.4cm$ is achieved through a combination of both strategies at the same time. } 
	\label{fig::hip_roll}
\end{figure}

As motivated earlier, a simple foot lift strategy through rotation of hip pitch, knee and ankle pitch joints brings the robot close to actuator velocity boundaries. However, we can use the hip and waist roll joints as well to lift the entire leg. This strategy is inspired by pelvic roll observations in human walking, and used by WABIAN-2 robot earlier \cite{hip_roll_webian_waist} to achieve straight-knee walking gaits. Figure \ref{fig::hip_roll} demonstrates both foot lift strategies side-by-side. In this work, we use a combination of the two to make sure that knee actuators stay far from their limits in normal conditions. This enables the actuators to reach fast yet feasible velocities in perturbed walking conditions. Lateral bounces in 3LP are as small as $2.2\ cm$ total displacement of the pelvis, given the desired step width of $25\ cm$ and step time of $T=0.4\ s$. These bounces induce sinusoidal trajectories of about $0.03\ rad \approx 1.7^o$ amplitude in the hip roll joints. However, a much larger pelvis roll of $0.1\ rad \approx 5.7^o$ leads to a foot lift of only $1.4\ cm$ (pelvis width is about $14\ cm$ in COMAN). Since the hip-roll strategy requires much larger motions in the joints, it becomes dominant, and there is no need to use lateral bounces of 3LP anymore. In practice, we command a sinusoidal pelvis roll trajectory of:
\begin{eqnarray}
\delta \theta_{roll}(t) = 0.1\ \sin(\frac{\pi t}{T})
\end{eqnarray}
and a half-sine vertical Cartesian trajectory of:
\begin{eqnarray}
\delta z_{fixed}(t) = 0.01\ \sin(\frac{\pi t}{T})
\end{eqnarray}
to the swing foot to lift it about $2.4\ cm$ in total. This makes the robot converge to a natural lateral bouncing, similar to WABIAN-2 robot \cite{hip_roll_webian_waist}. In this paper, we do not use any specific lateral footstep adjustment, momentum regulation or variable-timing phase-based control. Also, we set the ankle roll gain to a small value to let the foot adapt to the ground and establish at least a border contact. This increases the effective contact area in perturbed conditions \cite{collins2005bipedal}, provides more realistic ground reaction forces \cite{hamner2013rolling} and reduces unwanted transverse slippages during walking. 

\subsubsection{Sagittal Plane:}

Given a desired velocity, open-loop 3LP gaits $q(t)$ are generated in the Cartesian space and assigned to the robot through inverse kinematics. Both foot lift strategies augment the desired Cartesian positions and orientations of 3LP and given to our inverse kinematics controller \cite{faraji2017singularity}. We consider Cartesian pelvis and feet positions for tracking and state estimation in this work. The pelvis point is precisely in the middle of the two hip joints, and the foot position is on the ankle joint axis, slightly ($3\ cm$) shifted forward to be precisely in the middle of the foot. The advantage of considering the pelvis instead of CoM is avoiding the influence of upper-body arm oscillations which can highly perturb the CoM state. In COMAN, the two arms are rather heavy, and shoulder springs are much softer than those used in the legs. This can highly perturb the state estimation in perturbed walking conditions. 

\subsubsection{Position Control:}

In this work, similar to MIT's Spring Flamingo \cite{pratt1998intuitive} and many other robots, we use stance hip joints to regulate the torso angle instead of tracking the desired stance hip angle. This can be perfectly done with inverse dynamics and torque control and by using IMU orientations as feedback on pelvis orientations \cite{faraji2015practical}. In 3LP also, ideal stance hip actuators ensure an upright torso posture by calculating the required torque to be applied. Since we do not have torque control available in the present work, we only rely on position control. We use a simple time-based transition law that changes the desired and actual angles from swing command and hip encoder to zero torso angle and IMU pitch signal during swing and stance phases respectively. The hip actuator then tries to reduce the error by applying a proportional voltage. Denoting the IMU pitch by $\theta_{pitch}(t)$, the desired hip angle (in both phases) by $\theta_{des}(t)$ and the measured hip angle by $\theta_{act}(t)$, our transition law for the hip is formulated as:
\begin{eqnarray}
v(t) = k_d (\gamma(\theta_{des}(t)-\theta_{act}(t))+(1-\gamma)(-\theta_{pitch}(t)) 
\end{eqnarray}
where $\gamma = e^{-(t/t_1)^2}$ in stance phase and $\gamma = 1-e^{-(t/t_1)^2}$ in the swing phase, $t$ is the time since beginning of the phase and $t_1 = 0.2T$. The choice of $t_1$ is based on the force transfer rate during contact switch time (double support $\approx 20\%$), $v(t)$ is commanded actuator voltage and $k_d$ is the proportional gain. Figure \ref{fig::hybrid} demonstrates the switching rule for different moments of the gait. The position control for all other joints is simply defined by:
\begin{eqnarray}
v(t) = k_d (\theta_{des}(t)-\theta_{act}(t))
\end{eqnarray}
where $\theta_{des}(t)$ and $\theta_{act}(t)$ are the corresponding desired and actual angles. The control gains are set to $10\ V/rad$ for ankle roll joints, $100\ V/rad$ for hip pitch joints and $500\ V/rad$ for all other joints.

\subsection{Closed-Loop Control}

\begin{figure*}[]
	\centering
	\includegraphics[trim = 10mm 0mm 10mm 0mm, clip, width=1\textwidth]{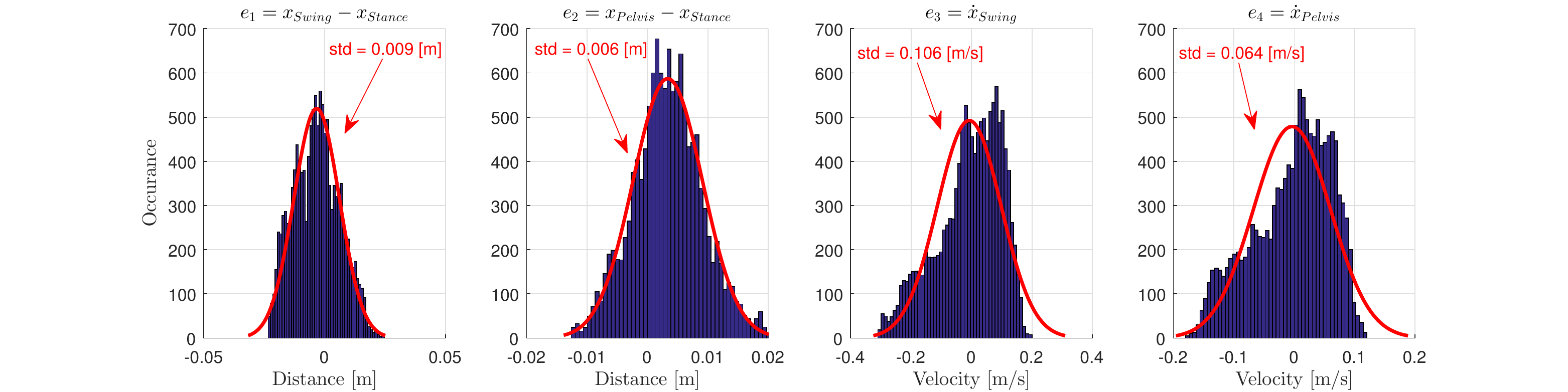}
	\caption{The quality of filtered state errors over a task of in-place walking. These values should be zero in ideal conditions. However, joint backlashes, calibrations and model mismatches in forward kinematics produce systematic errors that cannot be filtered out with a simple time filter of (\ref{eqn::time_filter}). We simply fit a Gaussian model on these measurements and estimate the typical standard deviation which is later used to determine the threshold for a dead-zone function.}  
	\label{fig::noise_reduction}
\end{figure*}

Our proposed control paradigm is based on planning and stabilization in the Cartesian space. We use inverse kinematics to convert Cartesian 3LP gaits to joint angles and then forward kinematics to calculate the actual Cartesian positions at each instance of time. In this section, we are going to discuss our closed-loop control which stabilizes the robot by dynamic footstep adjustment. The closed-loop controller has three main blocks: state measurement and filtering, adaptive lift strategy and footstep adjustment. 

\subsubsection{State Estimation:}

\begin{figure}[]
	\centering
	\includegraphics[trim = 0mm 0mm 0mm 0mm, clip, width=0.3\textwidth]{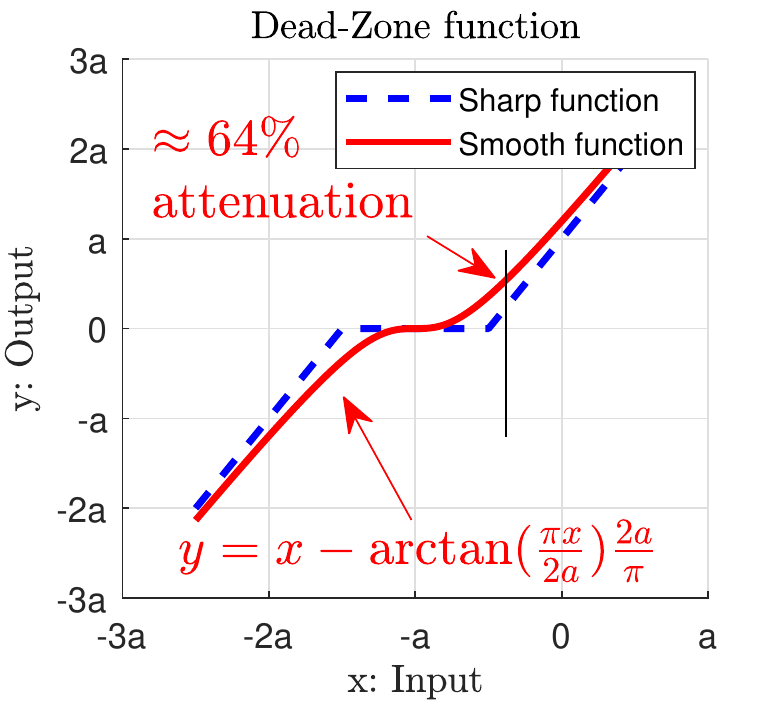}
	\caption{The smooth dead-zone function used to filter systematic errors due to backlashes, calibration and model mismatch problems. Compared to the sharp dead-zone function, the smooth function provides a considerable attenuation ($\approx 64\%$) which makes it suitable for error reduction.} 
	\label{fig::deadzone}
\end{figure}

We use link encoder (post-spring absolute encoder) and IMU measurements to build the kinematic chain and then to find relative horizontal pelvis-feet vectors $s_1(t)$, $s_2(t)$ (shown in Figure \ref{fig::3lp}) and their derivatives. Due to a quantization error, however, these vectors have a considerable noise to be filtered. Besides, joint backlashes and geometric model mismatches also contribute to a more systematic erroneous state measurement. We cancel the quantization error by a simple Infinite Impulse Response (IIR) filter of the form:
\begin{eqnarray}
s(t) = (1-\zeta) s(t-\delta t) + \zeta \hat{s}(t)
\label{eqn::time_filter}
\end{eqnarray}
where $\hat{s}(t)$ is the observed signal at time $t$, $\delta t \approx 2\ ms$ is our effective control rate and $\zeta=0.1$ is tuned to provide a right level of signal filtering and delay. In practice, this parameter was tuned together with position control gains on the hip pitch joints to ensure fast tracking and to avoid high frequency shakes due to actuation delays. To estimate velocities, we simply use time differentiation over the filtered $s_1(t)$ and $s_2(t)$ signals, and another averaging filter afterward with a window of $30$ samples. More systematic errors due to backlashes and model mismatches are canceled by introducing a dead-zone function. Consider the definition of instantaneous error vector $e(t)$ in (\ref{eqn::current_error}) which stacks errors of $s_1(t)$, $s_2(t)$ and their derivatives with respect to the nominal gait altogether (we call these errors $e_1(t)$ and $e_2(t)$ which are contained in $e(t)$ together with their derivatives). We run an open-loop in-place walking in which the robot only bounces left and right without any forward or backward progression. This gait is stable in a very limited region of states, depending on the stiffness of ankle joints and the feet size. Figure \ref{fig::noise_reduction} shows statistical values of filtered errors $e_1(t)$ and $e_2(t)$ and their derivatives in the sagittal direction. With ideal actuation and measurements, these values should be zero all the time. However, they have a certain standard deviation which we measure by fitting a Gaussian model on the data. A smooth dead-zone function:
\begin{eqnarray}
y = x - \arctan(\frac{\pi x}{2a}) \frac{2a}{\pi}
\end{eqnarray}
can attenuate the signal by $64\%$ at the threshold value $a$, shown in Figure \ref{fig::deadzone}. We apply such dead-zone on each of the signals in Figure \ref{fig::noise_reduction} and set the threshold $a$ to their standard deviation. Our two-stage filtering helps to stay in the limited basin of attraction produced by CoP modulation, but smoothly and automatically switching to foot-stepping strategy whenever disturbances get large. 

\subsubsection{Adaptive Foot Lift Strategy:}

\begin{figure*}[]
	\centering
	\includegraphics[trim = 0mm 0mm 0mm 0mm, clip, width=1\textwidth]{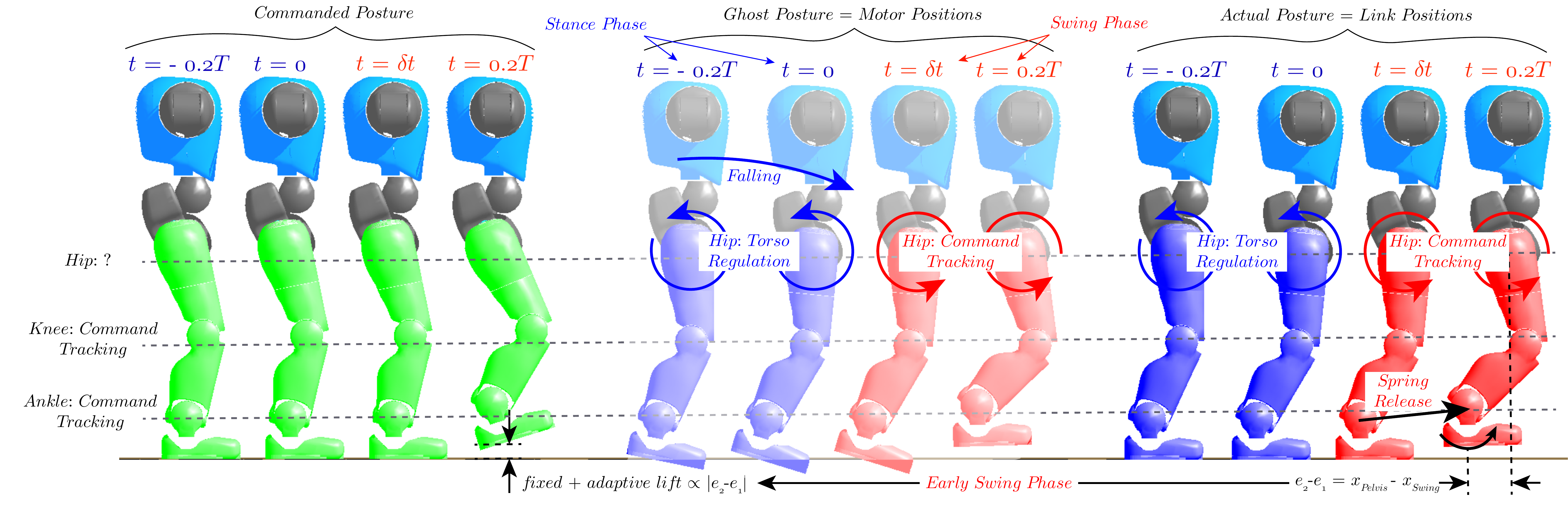}
	\caption{Phase transitions in commanded, ghost and actual postures defined in Figure \ref{fig::control}. The focus of this figure is to investigate hip switch rules and adaptive foot lift strategies, demonstrated here for example in an in-place walking gait. When the right leg is stance phase at time $t=-0.2T$, the commanded posture implies a pelvis position on top of the stance foot position. Due to external pushes, the robot might fall where the pelvis moves forward, getting away from the stance foot. This naturally happens since the stance hip is regulating the torso angle and not the desired hip angle. At time $t=0$, a phase transition happens while the robot is still falling. The toes of the ghost robot penetrate in the ground while ankle springs compress and move the CoP forward in an attempt to resist against the push passively. When the swing phase starts at $t=\delta t$, the actual hip angle is far from desired, the ankle spring is still compressed, and the toe in the ghost robot penetrates the ground. An adaptive foot lift strategy is therefore needed on top of our fixed strategies which are only designed for the commanded posture. The adaptive lift, depending on the current posture, provides an extra lift on the commanded posture which leads to enough ground clearance shown at $t=0.2T$. It also applies a corrective orientation to compensate leg rotation and ankle spring compressions (described better in Figure \ref{fig::ankle_deflection}).} 
	\label{fig::hybrid}
\end{figure*}

One of the key control components in perturbed walking conditions is foot lift which should provide enough ground clearance to allow for a complete swing motion and a precise tracking of next footstep locations, which then play important roles in stabilization. However, due to actuation limitations and especially velocity limits (shown in Figure \ref{fig::max_speed}), we cannot lift the foot in COMAN as much as desired. Previously, we introduced a mixture of knee-bending and hip-roll fixed foot lift strategies which already provide few centimeters of ground clearance. However, since the stance hip pitch is assigned to regulate the torso angle, the stance foot might get away from the pelvis, depending on external pushes and falling dynamics. Since in stance phase, we lose track of desired stance hip pitch trajectories in favor of torso regulation, the amount of leg lift needed in the early next swing phase should depend on the actual relative swing foot position. To better understand this effect, imagine the robot is performing in-place walking, and a strong push is applied from behind. The pelvis naturally moves forward with respect to the stance foot in the same phase while the torso is kept upright by the stance hip joint. In the next phase, the stance leg goes to swing mode, where the desired gait is only in-place walking. The desired swing hip angle is almost zero while the actual angle is such that this leg is left behind the robot in very early moments of the swing phase that has just started. Figure \ref{fig::hybrid} demonstrates this effect in details. At time $t=-0.2T$ for example, the right leg is in stance mode, and the robot is falling forward. Assume the phase transition happens at $t=0$ where the robot is still falling forward. At time $t=\delta t$, the right leg has started the swing motion, but its actual position is far from the commanded position. The fixed foot lift strategies are designed for the commanded posture (shown in Figure \ref{fig::hybrid}) while the actual swing foot position needs an extra lift to clear the ground safely. Our Adaptive foot lift strategy is straightforward. It calculates the relative horizontal pelvis-swing position out of the filtered error signals (i.e., $e_2(t)-e_1(t)$) and approximates the actual hip angle:
\begin{eqnarray}
\hat{\theta}_{act}(t) \approx \arctan(\frac{e_2(t)-e_1(t)}{z})
\end{eqnarray}
where $z$ is the constant pelvis height in 3LP model. This approximation is already filtered and registered in the inertial frame thanks to the IMU orientation while the actual post-spring hip encoder angle $\theta_{act}$ does not have these properties. The amount of adaptive leg left would be:
\begin{eqnarray}
\delta z_{adaptive}(t) = z(1-\cos(\hat{\theta}_{act}(t)))  \sin(\frac{\pi t}{T})
\end{eqnarray}
which compensates the mismatch between commanded and actual postures, shown in Figure \ref{fig::hybrid}. The use of a sinusoidal signal is to ensure extended legs at phase transition moments and to prevent the robot form collapsing in practice. At time $t=0.2T$, the adaptive lift strategy induces an extra lift in the commanded posture (shown in Figure \ref{fig::hybrid}) which goes to inverse kinematics and produces an extra foot lift, whenever the actual swing foot is unexpectedly far from the desired swing foot position.

\begin{figure}[]
	\centering
	\includegraphics[trim = 0mm 0mm 0mm 0mm, clip, width=0.5\textwidth]{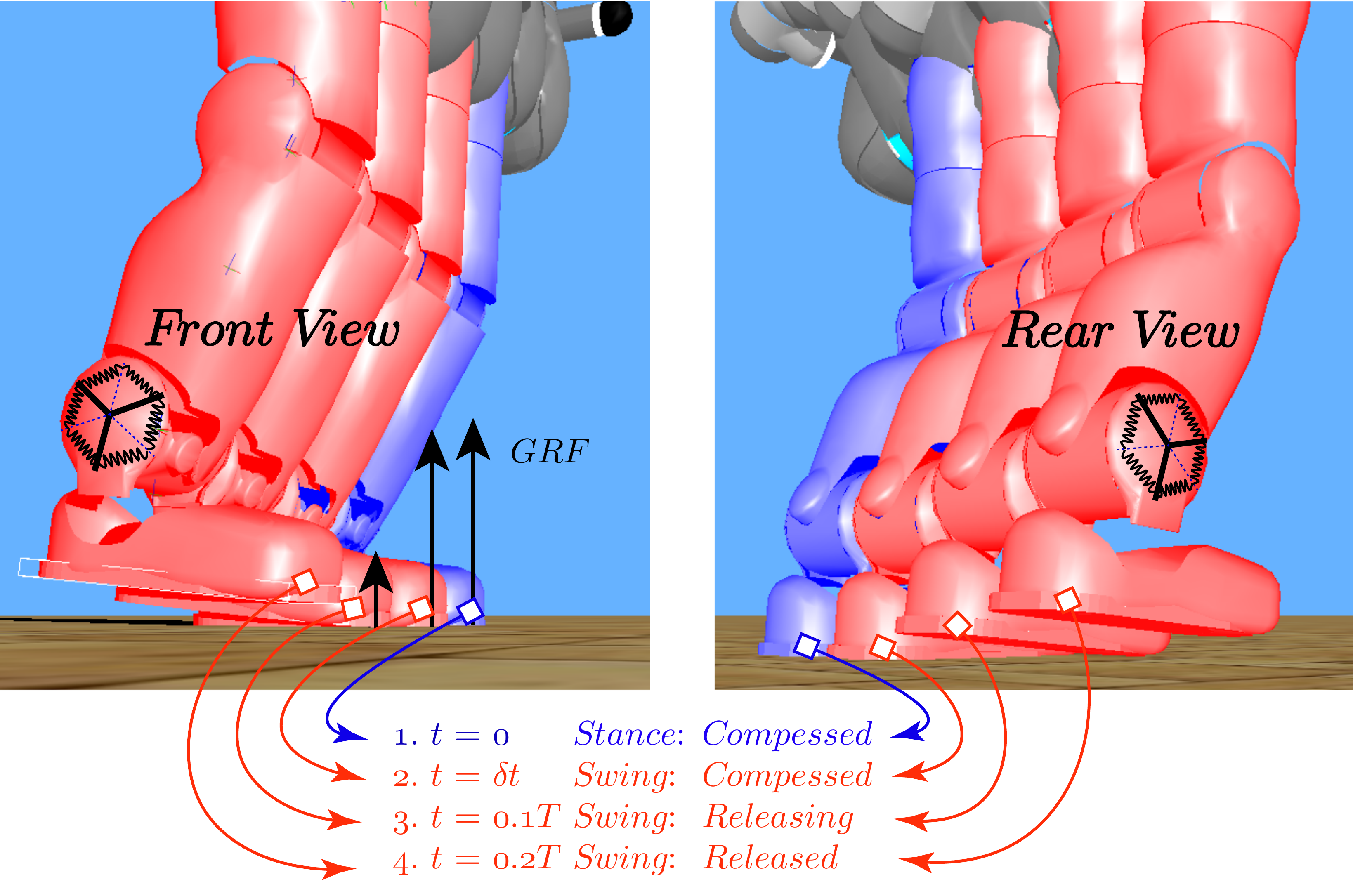}
	\caption{The effect of ankle spring compressions in early swing phases. Consider a phase change happens at $t=0$ where the stance leg switches to swing phase. Due to external pushes, the pelvis might move forward during the stance phase which brings the CoP forward too, since a fixed ankle angle is commanded. The movement of CoP is in fact due to the resistance of ankle spring against rolling which indeed provides positive stabilization properties, depending on the ankle spring stiffness. Now, when the swing phase starts at $t=\delta t$, the ankle spring is still compressed which continues to release during the leg lift. At some point, the spring releases completely which is when the actual toe-off happens, and the swing can start. In our adaptive lift strategy, we add some orientation correction feedback to speed up this effect.} 
	\label{fig::ankle_deflection}
\end{figure}

Due to hip switching rules and in addition to the foot height mismatches explained, the orientation of the foot might also not be horizontal in early swing phases (shown in Figure \ref{fig::hybrid}). This is purely due to rotation of the swing leg in the previous stance phase which has to be corrected in the current swing phase. This effect is shown in Figure \ref{fig::hybrid} on the ghost robot. Note that the commanded posture requires a flat foot, however in stance phase, the robot falls forward and the whole stance leg rotates, leaving the foot behind the pelvis. Now, since stiff position controllers are tracking the desired ankle angles, the ghost foot penetrates the ground. However, the SEA elements compress in this case and bring the CoP to the toes. This has a positive stabilization effect during the fall and slows it down. However, in the early next swing phase, although the robot attempts to lift the foot through fixed and adaptive strategies, the ankle spring is still releasing, causing the toes to keep touching the ground at time $t=0.1T$, shown in Figure \ref{fig::ankle_deflection}. Therefore, there are two distinct effects: I) foot orientation mismatch because of our hip switch rule, II) spring release which happens whenever the CoP goes towards the toes, regardless of the control algorithm. Figure \ref{fig::ankle_deflection} provides a closer perspective aiming at a better clarification of spring dynamics in the early swing phase. To compensate both effects, we simply add feedback on the foot orientation:
\begin{eqnarray}
\delta \theta_{pitch}(t) = - \hat{\theta}_{act}(t)
\end{eqnarray}
where $\hat{\theta}_{act}$ is coming out of the forward kinematics model. This feedback which is added to the desired foot orientation before inverse kinematics keeps the foot orientation always horizontal with respect to the ground. Therefore, our adaptive lift strategy is composed of both foot height and foot orientation compensations.

\subsubsection{Footstep Adjustment:}

\begin{figure}[]
	\centering
	\includegraphics[trim = 13mm 0mm 20mm 0mm, clip, width=0.5\textwidth]{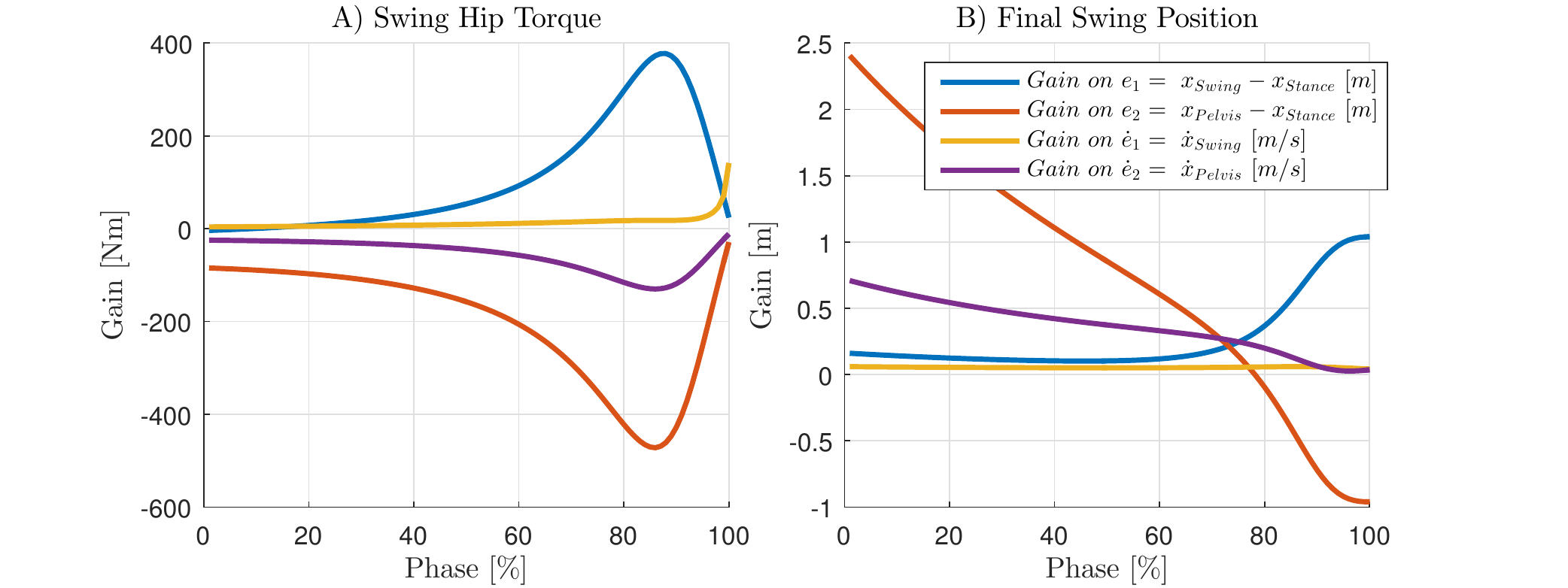}
	\caption{A) The instantaneous output of time-projecting controller on the 3LP model in terms of hip torques. B) The resulting footstep adjustment by simulating the current measured error until the end of the phase. In our position-controlled framework, we measure and filter errors at any time $t$, perform an inner product by the corresponding gain vector of these figures at time $t$ and truncate the result to find a safe final footstep adjustment. This value is then divided by the 3LP pelvis height $z$ to obtain an attack angle adjustment.} 
	\label{fig::projecting_gains}
\end{figure}

The second part of our closed-loop control is footstep adjustment based on time-projection control. Remember that the adaptive lift strategy was introduced to handle perturbed walking conditions and to make sure that desired footstep locations can be realized by providing enough ground clearance. After filtering errors $e_1(t)$, $e_2(t)$ and derivatives, we have an estimate of robot's current state error which is further passed through dead-zone functions to reduce systematic errors. This state variable is used in the time-projection controller to find footstep adjustments. In the absence of disturbances, the time projecting controller produces the same corrective input when the error evolves in time according to the natural system dynamics. In other words, once the external push is released, it leaves an error in the system which follows system's dynamical equations. In this regard, assuming short and bounded forces, we expect the time-projecting controller to produce the same final footstep in all time instances after the intermittent push until the end of the phase. Due to IMU and encoder noises, we have to decrease position controller gains in the hip joints to avoid shaking. This inevitable policy introduces more delay in the tracking of the desired hip trajectories, yet is enough to stabilize the system and realize commanded footstep locations. Because of these practical reasons and to set up a simple controller, we directly use the final footstep location to let the swing leg reach it on time. At each control tick (every $2\ ms$ approximately), we take the current Cartesian state and simulate it until the end of the phase in a closed-loop 3LP simulation where the time-projecting controller adjusts the swing hip torques. The resulting footstep adjustment at the end is extracted (from posture $F$ in Figure \ref{fig::control}) and converted to a proper attack angle.

In practice, thanks to the linearity of 3LP and time-projection control, we can simply extract control laws off-line and interpolate them in the form of a look-up-table on-line which is computationally very efficient. Note that the effective sagittal error vector used here has 4 dimensions (sagittal components of $e_1(t)$, $e_2(t)$ and their derivatives). The output of projecting controller has two dimensions in the form of constant and time-increasing torques which are added together at each instance of time $t$ and produce a single swing hip torque values. On the other hand, when simulating the current error until the end of the phase, the time projecting controller produces a footstep adjustment in the sagittal direction which has one dimension. Therefore, two different look-up-tables can be produced, one for a torque-controlled paradigm and one for our position-controlled paradigm. Each control table maps a four-dimensional error into a single output. The control gains of individual error dimensions are shown in Figure \ref{fig::projecting_gains} at different phase times. These gains depend on model properties and the walking gait frequency, but not the speed \cite{faraji20173lp2}. In practice, we only take filtered errors and perform inner product with an interpolation of the final swing position control gains in Figure \ref{fig::projecting_gains}. A truncation of $15\ cm \approx 0.35 z$ is also applied to avoid huge steps and breaking the hardware when too strong pushes lead to failure during experiments. Results are finally divided by the 3LP pelvis height $z$ to approximate the equivalent attack angle adjustment.

Having explained both open-loop and closed-loop simple control policies, we are interested in assessing the effect of each control block as well as push recovery and walking scenarios in different conditions. These results are presented in the next section.

\section{Results}

The results presented in this section are categorized into five groups of in-place walking, intermittent push recovery, continuous push, normal walking and stiffness measurement scenarios. We explore the functionality of closed-loop and open-loop control configurations where the foot-stepping mechanism is just disabled in the second case. We further continue with the closed-loop controller and in-place walking and show how the algorithm can take corrective steps to recover pushes in different directions. Next, we disable the footstep adjustment and demonstrate the limited stability of open-loop controller which is due to passive CoP modulation. This controller can easily follow a continuous external push and pull while behaving almost on the margin of stability. Finally, we show how different walking gaits can be generated with this framework where we only change the desired speed in the 3LP gaits. Finally, we perform a simple standing experiment and demonstrate the effective level of ankle stiffness in COMAN which further clarifies dynamic walking features of our controller. 

\subsection{In-Place Walking}

\begin{figure}[]
	\centering
	\includegraphics[trim = 0mm 0mm 0mm 0mm, clip, width=0.5\textwidth]{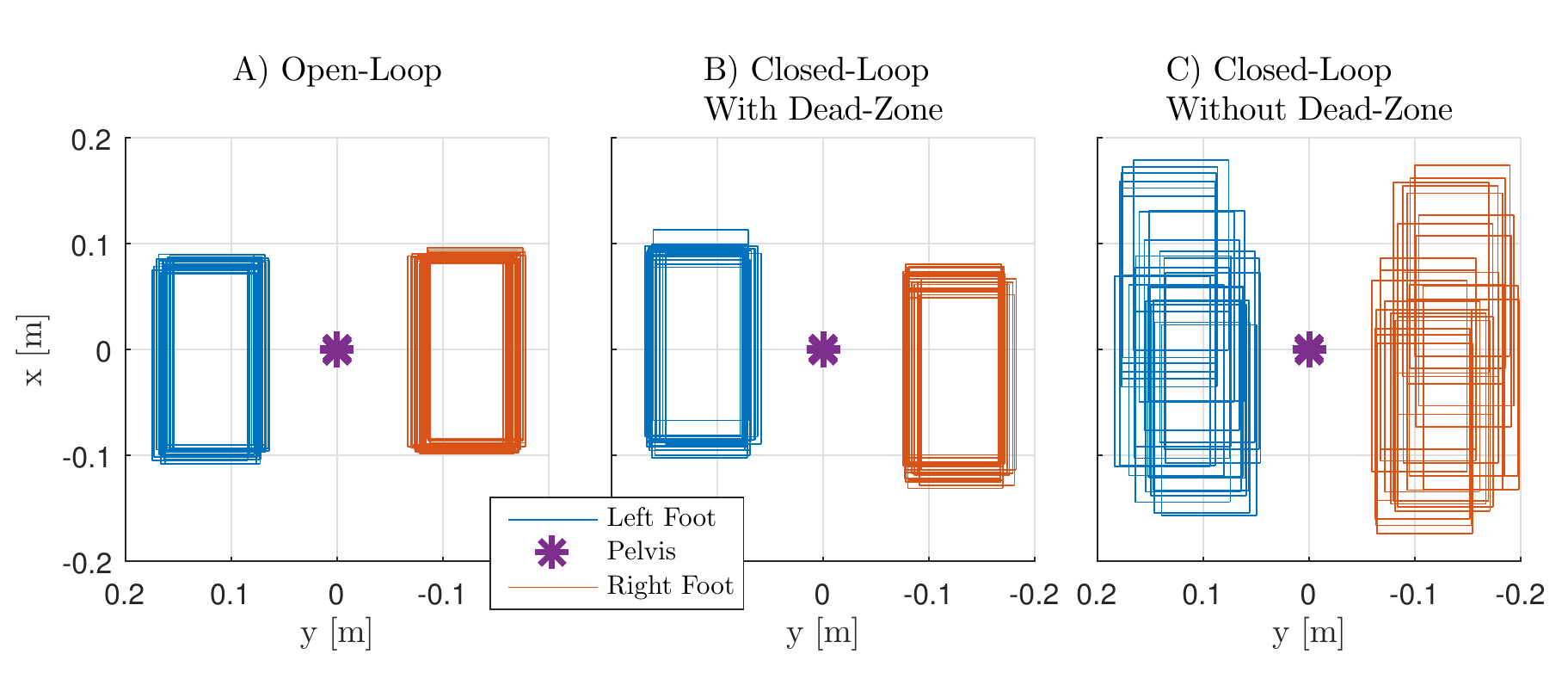}
	\caption{In-place walking in A) open-loop, B) closed-loop and C) closed-loop without dead-zone functions. This figure shows horizontal foot locations only at phase change moments over $10\ s$ of walking, i.e., $25$ steps. The open-loop gait is stable in a very limited region of states thanks to the passive CoP modulation of ankle springs and the natural swing trajectory which always comes under the pelvis during an in-place walking gait. When applying dead-zone functions in closed-loop control, the in-place walking gait is only slightly perturbed in figure B). Without the dead-zone functions, however, the walking gait is completely perturbed. Here, the footstep still stabilizes the system, but the timings and trajectories are all perturbed systematically, which is not desired. The corresponding videos of all the three scenarios could be found in Multimedia Extension 1.} 
	\label{fig::inplace}
\end{figure}

As motivated in the beginning, CoP modulation provides a fast yet limited control authority for immediate stabilization. Foot-stepping, on the other hand, provides stronger stabilization, but only over phase changes. SEA elements and internal joint backlashes lead to a relatively compliant ankle joint, shown at the end of this section. The passive ability of ankle springs in resisting against falling together with the fact that during in-place walking, the swing foot comes under the pelvis all the time make the open-loop gait stable in a limited region of states without any time-projection control. Figure \ref{fig::inplace}.A shows foot positions with respect to the pelvis during this open-loop gait which is also used in Figure \ref{fig::noise_reduction} to design the dead-zone functions. In Figure \ref{fig::inplace}.B, we show how the complete closed-loop system behaves when dead-zone functions reduce systematic errors of Figure \ref{fig::noise_reduction}. 

\subsubsection{Without Dead-Zone:}

\begin{figure}[]
	\centering
	\includegraphics[trim = 0mm 0mm 0mm 0mm, clip, width=0.45\textwidth]{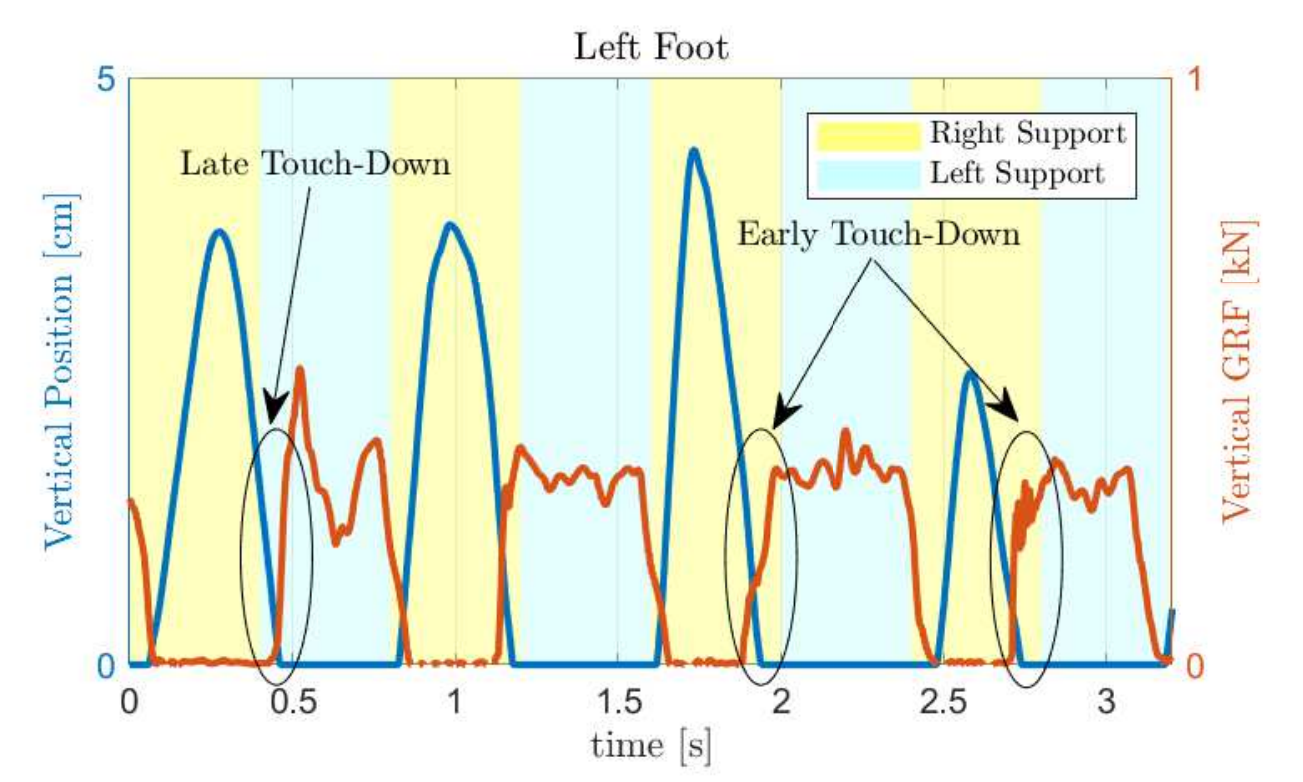}
	\caption{ Closed-loop in-place walking when the dead-zone functions are disabled. In this case, the robot stumbles considerably due to the systematic errors in state estimation. The time-projecting controller stabilizes the system despite an erroneous measured state. However, actual phase change moments might happen earlier or later, depending on adaptive foot lift trajectories which are also subject to an erroneous measured state. This illustrates the effectiveness of dead-zone functions in improving the repeatability of closed-loop in-place walking gaits.} 
	\label{fig::timing}
\end{figure}

\begin{figure*}[]
	\centering
	\includegraphics[trim = 0mm 70mm 0mm 70mm, clip, width=1\textwidth]{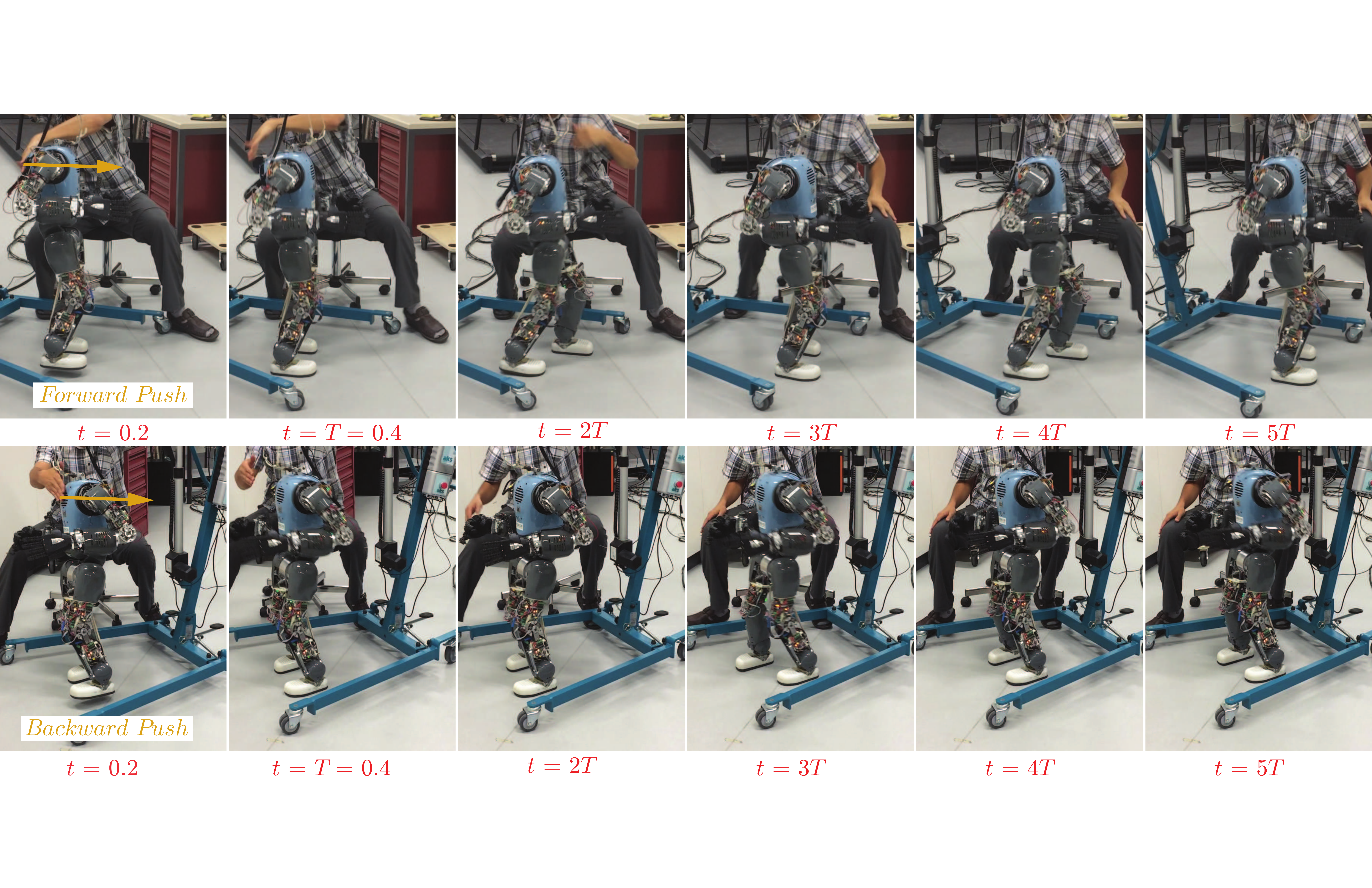}
	\caption{Short intermittent pushes during an in-place walking gait in different directions. On top, the robot almost takes three corrective steps to recover the forward push while it only takes two steps the recover a lighter backward push. Snapshots are taken at phase change moments every $T=0.4\ s$. The corresponding videos could be found in Multimedia Extension 1.} 
	\label{fig::push}
\end{figure*}

\begin{figure*}[]
	\centering
	\includegraphics[trim = 0mm 130mm 0mm 130mm, clip, width=1\textwidth]{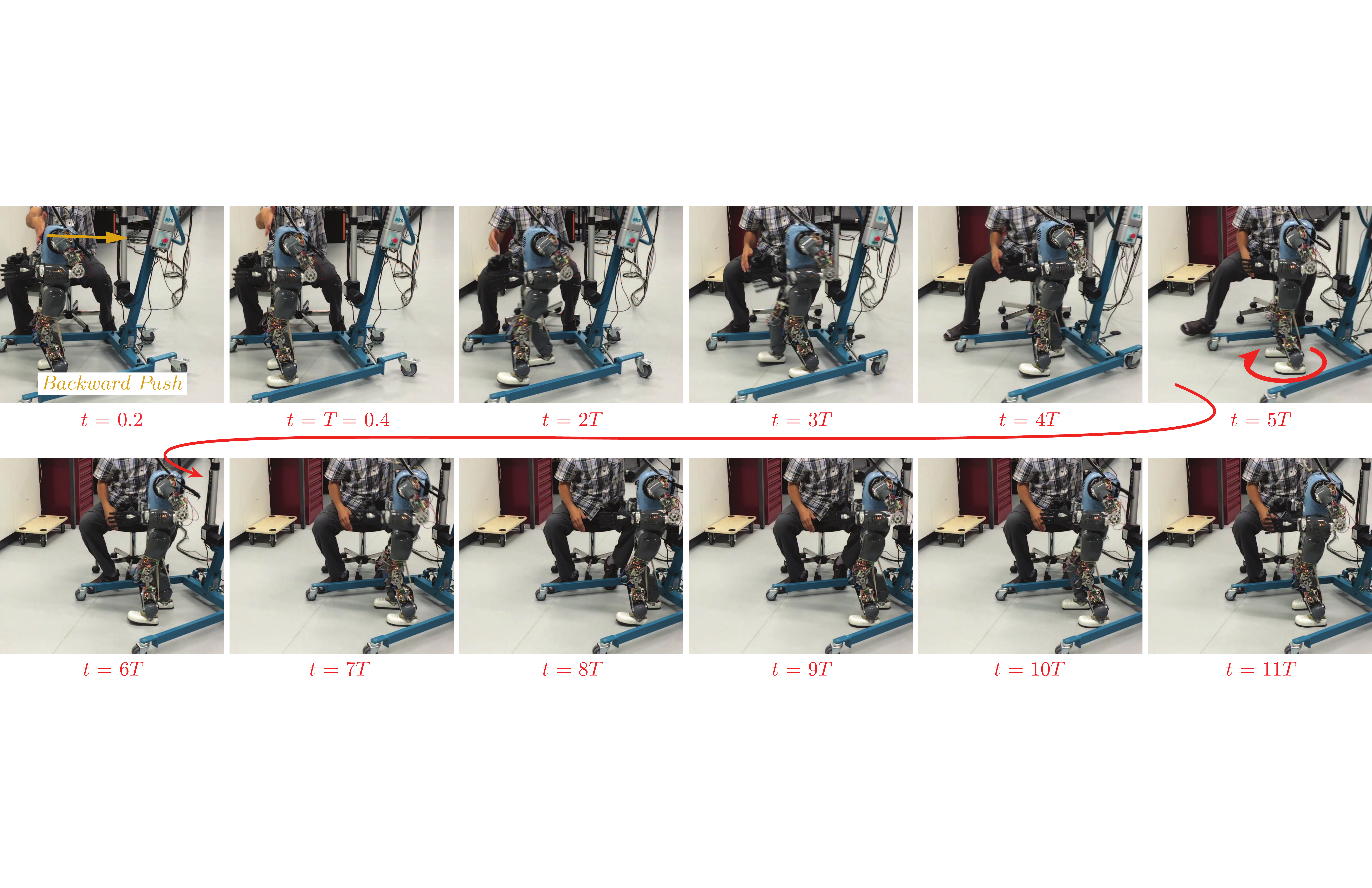}
	\caption{Recovery process of a strong backward push during in-place walking. Snapshots are taken at phase change moments every $T=0.4\ s$. Since the passive CoP modulation is almost ineffective in the backward direction, the stepping strategy takes multiple steps to recover the strong push. Despite a safety threshold which limits the step size, the push is so strong that it violates decoupling assumptions, leading to a sudden rotation at $t=5T$. The new undesired perturbation requires the robot to take more backward steps to recover completely and go to a rest condition. The corresponding video could be found in Multimedia Extension 1.} 
	\label{fig::strong_push}
\end{figure*}

Without dead-zone functions, the closed-loop system gets perturbed completely. This is shown in Figure \ref{fig::inplace}.C where the robot is still stable, but stumbling considerably. In these conditions, the lateral stability of the robot is also systematically influenced because of unexpected touchdown events shown in Figure \ref{fig::timing}. Here, the adaptive foot lift strategy commands different clearance heights due to erroneous measured states. This can lead to unexpected touchdown events which prevent a complete swing when happening earlier than expected or produce an early stance phase when the leg is still in a swing phase. 

\subsection{Intermittent Push Recovery}

Although the passive CoP modulation provides little stability over in-place walking in the absence of considerable disturbances, we would like to know how effective the foot-placement strategy is. We use the same closed-loop in-place walking gait of the previous part and apply short intermittent pushes on the torso.

\subsubsection{Moderate Pushes:}

Figure \ref{fig::push} shows a sequence of walking snapshots taken at phase change moments when a forward and backward push is applied to the system. Depending on the timing of the push, a small footstep adjustment takes place in the same phase with the push, but the next footstep locations recover the push gradually. Due to the safety threshold ($=15\ cm$) applied on footstep adjustments and depending on the push strength, the recovery process might take more than $1-2$ steps. Note that we typically command the pelvis in the middle of the support polygon. Ankle SEA springs, therefore, compress slightly as shown in Figure \ref{fig::ghost} due to a non-zero ankle torque when the CoP is not exactly under the ankle joint. Because of this default compression, the passive CoP modulation strategy resists more against forward pushes than backward pushes. Although most of the stabilization comes from the stepping strategy, overall, recoverable forward pushes can be slightly stronger than backward pushes.

\subsubsection{Strong Pushes:}

Our control design and safety thresholds target small, yet dynamic footstep adjustments which do not violate the linearity of 3LP, sagittal-lateral decoupling, and actuator limitations. When a strong push is applied, some of these assumptions might not be valid anymore. The robot might be still able to recover, but over multiple steps or with an extra effort. Figure \ref{fig::strong_push} shows such strong backward push during in-place walking. As mentioned earlier, the passive CoP modulation is almost ineffective in the backward direction which further complicates the recovery for the stepping strategy. The robot takes multiple backward steps to recover the strong push, but suddenly, a transverse slippage happens in the right foot which rotates the robot slightly. This happens due to the violation of sagittal-lateral decoupling assumptions. After this slippage, the robot takes a few other backward steps to stabilize completely.

\subsection{Continuous Pushes}

\begin{figure*}[]
	\centering
	\includegraphics[trim = 0mm 0mm 0mm 0mm, clip, width=1\textwidth]{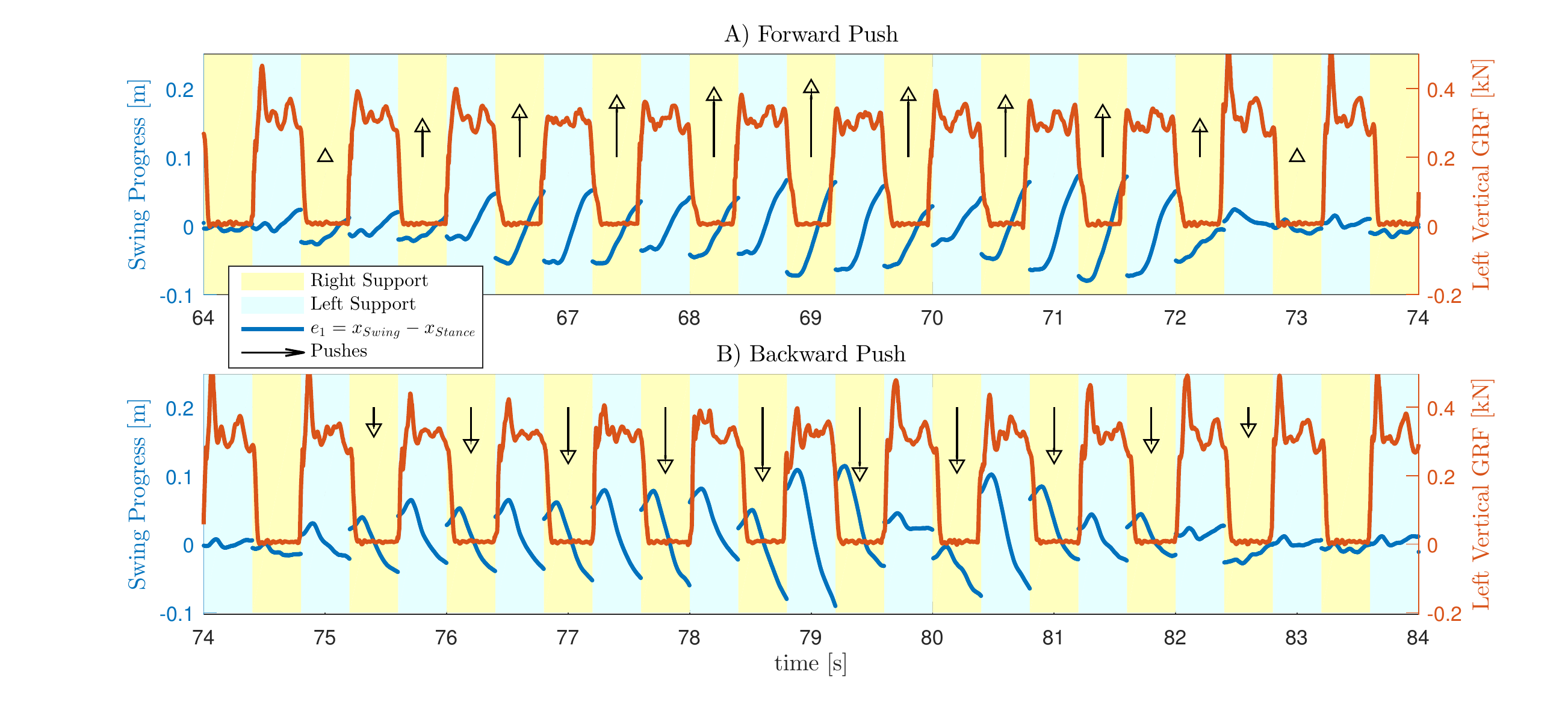}
	\caption{The emergent walking of the open-loop controller when an in-place walking gait is commanded and a moderate yet continuous external A) forward and B) backward force is applied. In-place gait requires swing foot to come under the pelvis while the external drag causes a continuous falling. Passive CoP modulations in the ankle SEAs slightly resist the force and provide a weak stability. However, the gait does not have a large basin of attraction and requires a smooth profile of external forces. The important property of this controller is compliance to external pushes and emergent speed modulations, although no specific desired velocity is commanded. In this scenario, the robot can sometimes reach a peak velocity of $\pm0.2\ m/s$. The corresponding video could be found in Multimedia Extension 1.} 
	\label{fig::following}
\end{figure*}

Apart from impulsive pushes, we are also interested in investigating the performance when the robot is subject to a moderate continuous pushing/pulling force. In these conditions, bounded-push assumptions behind the time-projecting controller are violated. As mentioned earlier, the open-loop controller is also stable due to a passive CoP modulation, but in a very limited region of states. This controller weakly reacts to perturbations, almost on the margin of stability which makes it compliant in following the pushing/pulling force. Figure \ref{fig::following} demonstrates this behavior in the presence of continuous pulling and pushing forces applied to the right hand. In these scenarios, we disable the stepping strategy while still keeping the adaptive foot lift active. Depending on actuator capabilities, the robot can easily reach walking speeds of up to $\pm0.2\ m/s$. When enabling the stepping strategy, however, the robot naturally uses footstep adjustments to slow further down. This leads to a stumbling behavior and a less rhythmic motion compared to the open-loop controller. Movies of both scenarios are included in Multimedia Extension 1.

\subsection{Walking}

\begin{figure}[]
	\centering
	\includegraphics[trim = 0mm 5mm 0mm -4mm, clip, width=0.5\textwidth]{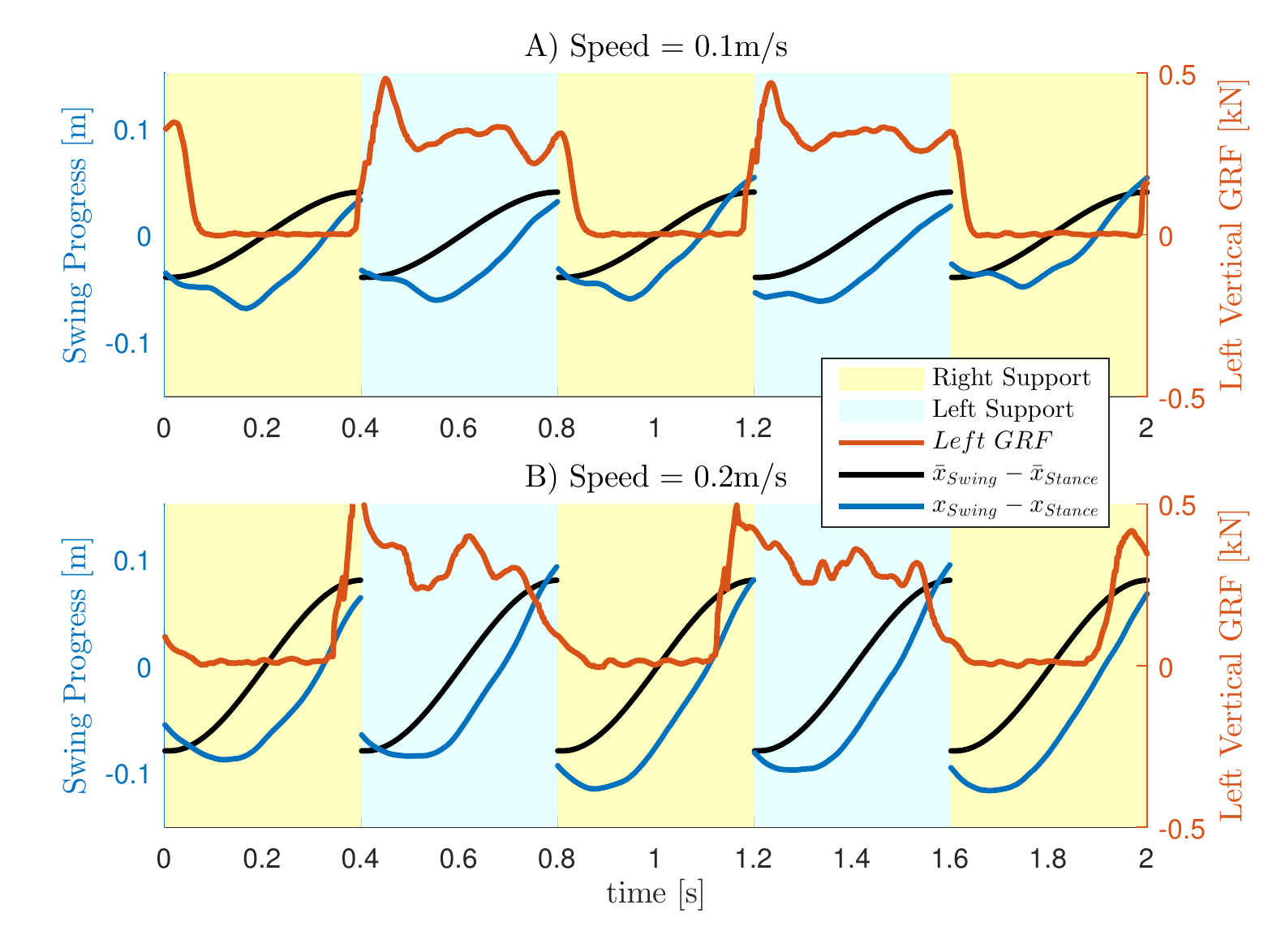} \\
	\includegraphics[trim = -30mm 20mm 0mm 10mm, clip, width=0.42\textwidth]{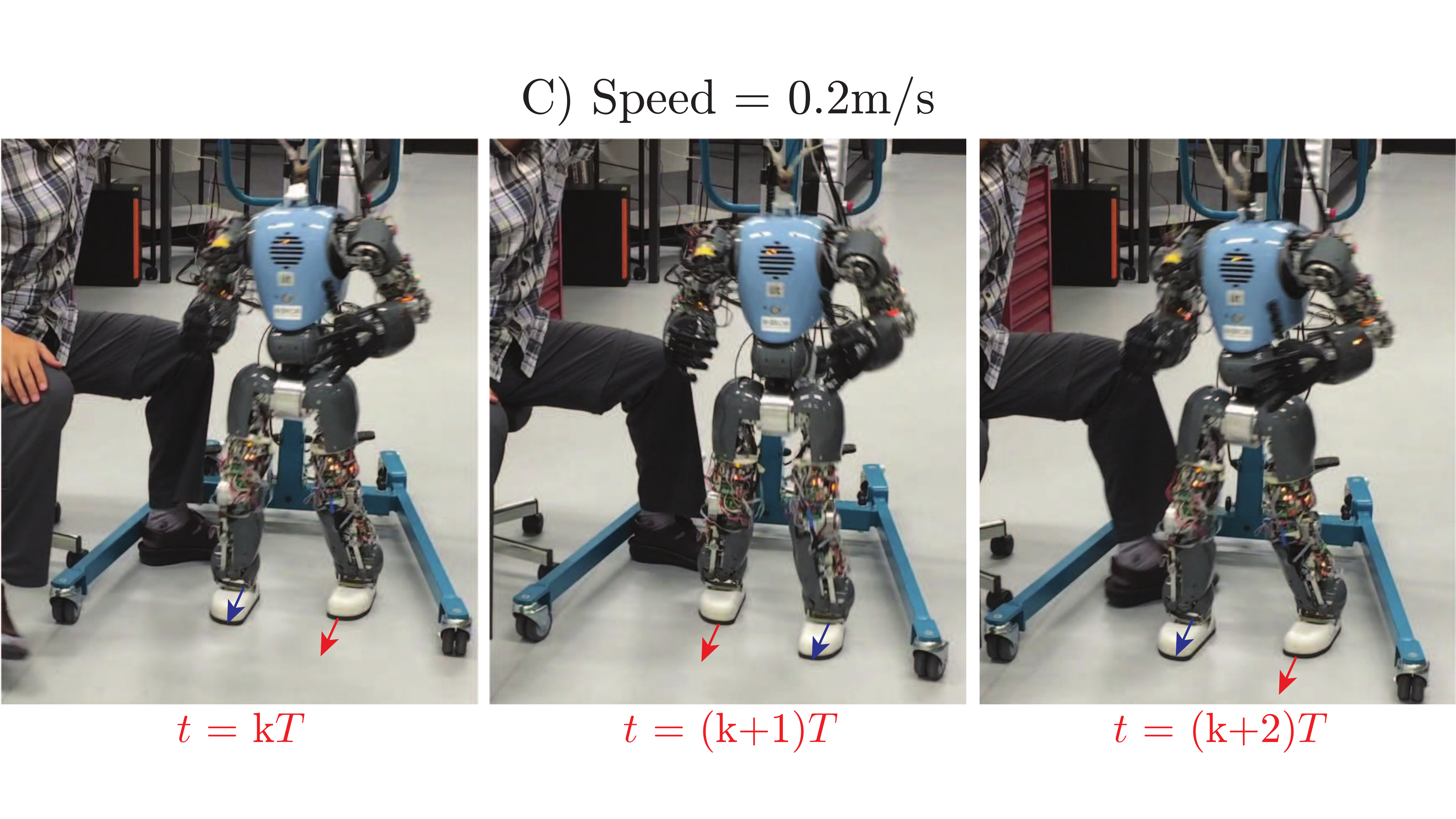}
	\caption{Walking gait generation with the same controller used previously for in-place walking. In these scenarios, we command a desired velocity of A) $0.1\ m/s$ and B) $0.2\ m/s$ which produces different 3LP gaits and Cartesian profiles. Here, the passive CoP modulation has little stabilization role while the time-projecting controller plays an important role, observed in small trajectory variations. Figure C) shows snapshots of the faster gait on the real robot, taken at phase transition moments. Movies of both scenarios could be found in Multimedia Extension 1.} 
	\label{fig::walk}
\end{figure}

Remember that due to the linear properties of 3LP and the time-projecting controller, with a fixed walking frequency, the controller remains unchanged at different walking speeds. While keeping the same closed-loop control architecture, we now change the desired speed in 3LP to produce variable trajectories. Figure \ref{fig::walk} shows two different cases of walking at $0.1\ m/s$ and $0.2\ m/s$ together with snapshots from the robot. The same dynamic stabilization mechanism can be seen in small variations of swing foot trajectories where the controller provides adjustments for stabilizing the gait. One can notice that, however, the actual phase transition takes place slightly earlier than expected in faster walking speeds. This can be due to a tracking delay in the hip joints or extra spring compressions in the stance leg which limit the ground clearance and influence the swing phase timing. The time-projecting controller can still stabilize the system, handling these timing issues.

\subsection{Effective Ankle Stiffness}

\begin{figure}[]
	\centering
	\includegraphics[trim = 0mm 0mm 0mm 0mm, clip, width=0.5\textwidth]{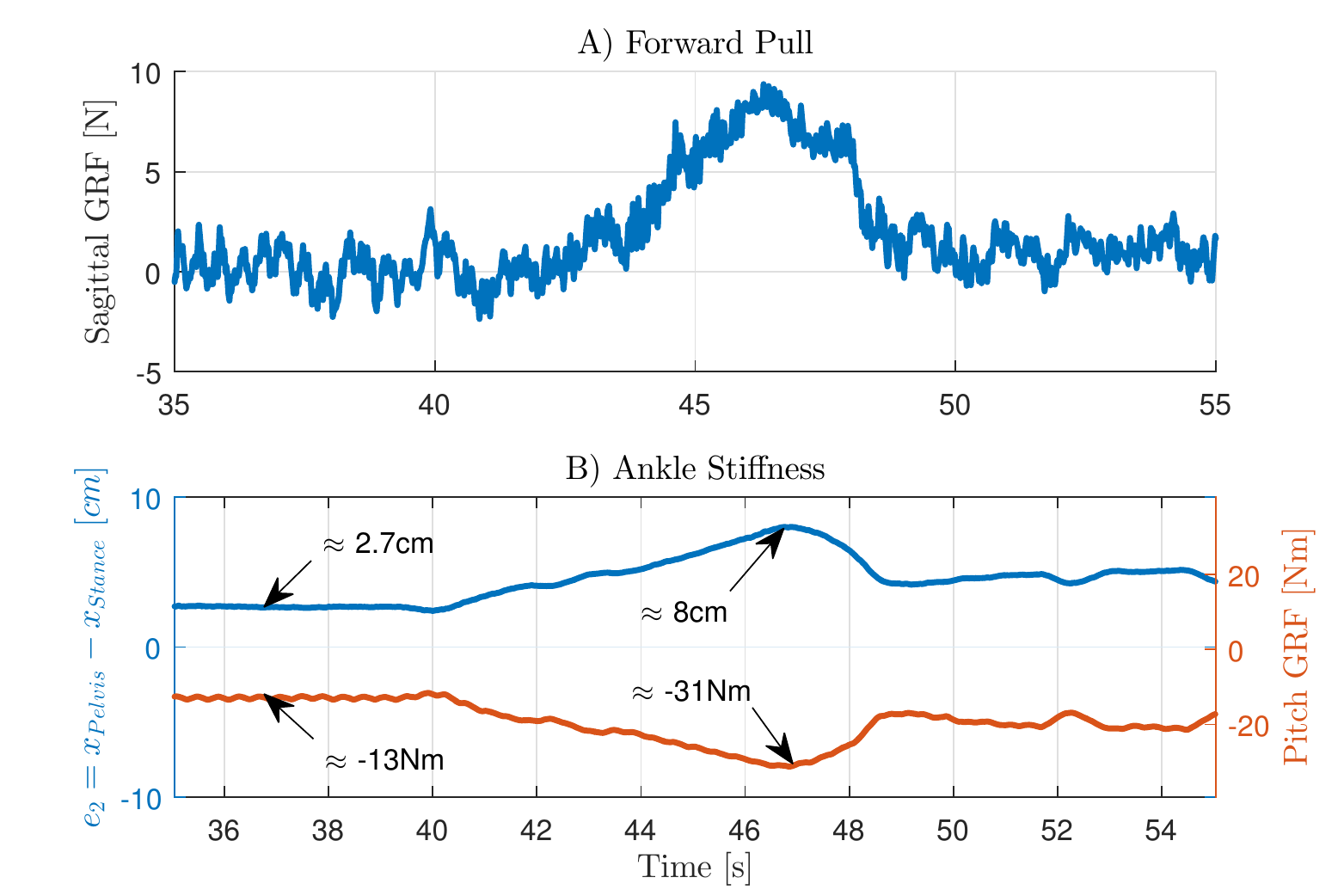} \\
	\includegraphics[trim = -25mm 10mm 0mm 0mm, clip, width=0.42\textwidth]{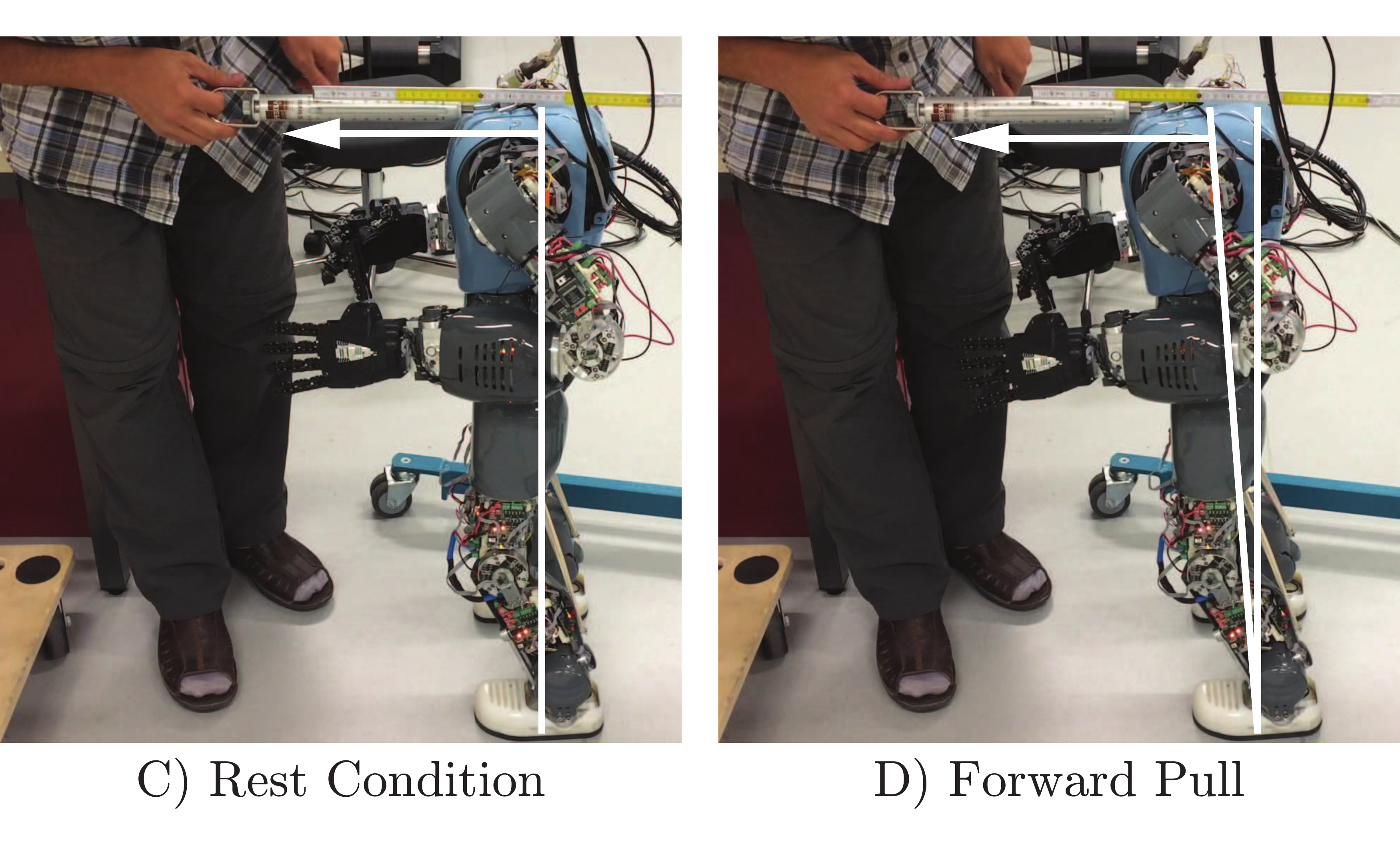}
	\caption{Standing robot subject to external pulling forces applied to the neck. A) Filtered sagittal forces measured by contact force sensors of the robot. B) Measured contact moments and the pelvis displacement reported by forward kinematics. C) The actual rest condition. D) The maximum forward lean due to the external pulling force. One can observe that with a minimal external force, the pelvis in COMAN can easily move forward which induces an effective stiffness of $\approx 31\ Nm/rad$ in each ankle spring. The virtual compliance realized by the position controller on the motor shaft is much stiffer, effectively negligible in series with such compliant physical springs. The corresponding video could be found in Multimedia Extension 1.} 
	\label{fig::stance}
\end{figure}

The last experiment presented in this section aims at quantifying the actual passive stiffness of ankle springs in COMAN. To this end, we command a stand-still posture to the robot with stiffest position gains everywhere, the CoP in the middle of support polygon and disabled feedbacks. In this scenario, as depicted in Figure \ref{fig::stance}, both ankles contribute a resisting force against external pushes applied to the neck. As Figure \ref{fig::stance}.A and Figure \ref{fig::stance}.D show contact force readings and external gauge values respectively, a minimal pulling force of $\approx 7N$ can easily move the pelvis by $\approx 5cm$ (and the torso by $\approx 9cm$) which implies an effective stiffness of only $\approx 31\ Nm/rad$ per ankle. Note, however, that the position controller with a high gain of $k_d=500$ and an effective voltage-torque coefficient of $1/\beta \approx 2.2$ leads to a stiffness of $\approx 1100\ Nm/rad$ in series which is much larger than the overall stiffness. Besides, we also have small backlash problems in the joints which further limit the influence of ankle springs. This experiment illustrates that, despite having relatively large feet in COMAN, our controller is mostly relying on the footstep adjustment strategy rather than the passive CoP modulation in recovery from disturbances.

\section{Conclusion}

In this paper, we proposed a very simple but robust control architecture that can stabilize the robot against external pushes. Motivated by specific properties of our hardware, including heavy legs, we used a model called 3LP which can capture swing and torso dynamics and produce more dynamically consistent gaits compared to LIP. We also used the rich literature of Poincar\'e maps and DLQR controllers for walking control together with our time-projecting controller that resolves the continuous-discrete mapping and offers an on-line control scheme. This controller, in a very simple position-controlled architecture, can suggest footstep adjustments that stabilize the robot against external disturbances. By using the 3LP model, we are performing a model-based control. However, thanks to a close dynamical match of this model to the real robot, both gait generation and on-line control in our architecture take falling and swing dynamics into account. This knowledge is encoded in a very simple control look-up-table that stabilizes the robot easily without using any advanced state estimation algorithm like Kalman filtering or advanced inverse dynamics.

\subsection{Dynamic Walking}

Over an extensive range of results and analysis, we showed that both passive CoP modulation and foot-stepping strategies are involved in our controller. The CoP modulation alone has a limited basin of attraction while the foot-stepping strategy is sensitive to the sensory data but more powerful in recovering strong disturbances. We resolved the sensitivity with simple filters and dead-zone functions on the sensory data. However, the effective ankle stiffness was determined by the physical hardware and much smaller compared to the virtual compliance realized through position controllers. Our architecture is not designed for an active CoP or Zero Moment Point (ZMP) control, and we only rely on the passivity of ankle springs to resist or at least damp falling dynamics. CoP modulation has, therefore, a very limited influence regarding COMAN's foot size and effective ankle stiffnesses. Our controller provides stability against very small disturbances through a passive CoP modulation and recovers more substantial disturbances dominantly through an active foot-stepping strategy. Our proposed algorithm can be suitable for taller robots like Atlas or robots with even smaller feet sizes like Marlo, Cassie or Humo as well. This is an advantage for our model-based controller which encodes robot and gait properties all in the 3LP model. It produces then a consistent time-projection look-up-table control which only depends on the gait frequency. Position control gains and filtering constants are simply chosen to provide the stiffest control without shaking which could be tuned for other robots easily during a simple standing task.

\subsection{Decoupling Assumptions}

We focused on sagittal plane dynamics only while producing lateral bounces through simple open-loop foot lift strategies existing in the literature. The relatively wide choice of step width improved the lateral stability, though limited the sagittal-lateral decoupling. Our controller is nevertheless limited to the constant-height assumption of 3LP as well which further limits step sizes in practice. Besides, velocity limitations of actuators also do not allow for fast swing motions. Inspired by numerous works on simple passive walkers with wide feet and human observations, we decreased the ankle roll stiffness and allowed for a wider foot contact in practice which increases available transverse torques and prevents unwanted turning around the yaw axis. For most of the experimental scenarios presented in this paper, the decoupling assumption was valid. When applying large pushes, however, the robot takes longer steps which might lead to small rotations, yet capturable with our powerful foot-stepping algorithm.

\subsection{Constant CoM height}

This assumption is initially used in the 3LP model to derive linear equations of motion which computationally facilitate gait generation and control. The idea of time projection can be extended to nonlinear models too, as long as one can linearize the system around a certain periodic trajectory and find any-time transition matrices $A(t)$ and $B(t)$, used in time-projection. We believe that the simple control look-up-tables found here can be used in other robots as well through the adjustment of model properties and the desired walking frequency. All the knowledge of falling and swing dynamics together with the optimality of LQR design are encoded in these look-up-tables which produce very simple swing hip torques or desired attack angles. A fixed frequency is an underlying assumption behind 3LP and time-projection. However, we showed that the architecture could tolerate significant phase mismatches too. This makes time-projection suitable for phase-based bipedal walkers and robots taller than COMAN. The algorithm might also work on smaller robots. However, they usually have larger feet which make them statically very stable.

\subsection{Hardware Limitations}

The present manuscript features discussions of COMAN hardware in a very deep level. We presented actuator models, velocity limits, sensory data qualities, actuation delays, backlashes and systematic model errors. We used a mixture of fixed foot lift strategies to maximize the lift and to avoid overloading actuators in terms of velocities. The dynamic matching of 3LP with the robot is so close that our controller does not need a good quality of state estimation and control. We only read encoder and IMU positions while relying on internal actuator properties for a high-frequency damping. We also use very simple filtering policies that are all tuned based on actual robot data, position control gains, and control loop delays. The unified control framework has only a few parameters to tune and does not include tunning of any critical parameter. A higher amount of foot lift, of course, pushes actuators towards their velocity limits but improves the stability. A stronger filtering provides cleaner signals which allow for higher gains in the hip joints, but in practice increase the delay too which is not desirable in our relatively high walking frequency. 

\subsection{Compliance}

Our control architecture includes specific components to compensate for systematic errors arising from hardware compliance, but not fighting against this advantageous property that protects the hardware in perturbed walking conditions. We used a simple compensation in the knee springs to adjust unwanted rotations and pelvis shifts. We also used a simple orientation feedback to compensate ankle spring deflections in early swing phases. These simple policies improve stability considerably. We also employed a switching control rule for the hip joints which is also inspired by the vast body of literature on planar walking robots. This policy regulates the torso orientation while providing natural falling dynamics which is originally encoded in the stance hip joints of 3LP too. We believe a torque control paradigm can further increase compliance, but the available level of compliance in the physical SEAs of the robot was enough for our perturbed walking conditions. 

\subsection{Future Work}

In future, we would like to extend this framework to the lateral direction too by finding lateral foot-placement suggestions. Since dynamic equations are the same in 3LP for both sagittal and lateral planes, the same control look-up-tables can be applied in the lateral direction. However, one should carefully handle internal collisions which might happen in case of large lateral pushes. We hope to reduce the step width by an active control of lateral stability which can then help reducing the overall walking frequency and provide a more human-like walking. We also aim at upgrading our low-level joint controllers with feed-forward velocity and torque terms calculated by 3LP and the time-projecting controller. This can improve compliance and tracking delays. Finally, we consider improving low-level control-board issues, renewing springs and improving backlashes on COMAN to increase the voltage, to reduce mismatches and to improve sensory precision respectively. This can further enlarge the operational region of actuators and the overall controller. Hardware imperfections and possible model mismatches always exist, but dynamic foot placement is strong enough to stabilize the robot through hybrid walking phases. This motivates us to add more human-like features like CoM excursions, straight knees, and toe-off phases to achieve faster speeds in future work. This paper is accompanied with a Multimedia Extension demonstrating walking and push recovery scenarios. All codes used in this paper are available online (after being accepted) at \url{http://biorob.epfl.ch/page-99800-en.html}.

%%%%%%%%%%%%%%%%%%%%%%%%%%%%%%%%%%%%%%%%%%%%%%%%%%%%%%%%%%%%%%%%%%%%%%%%%%
%%%%%%%%%%%%%%%%%%%%%%%%%%%%%%%%%%%%%%%%%%%%%%%%%%%%%%%%%%%%%%%%%%%%%%%%%%
%%%%%%%%%%%%%%%%%%%%%%%%%%%%%%%%%%%%%%%%%%%%%%%%%%%%%%%%%%%%%%%%%%%%%%%%%%
%%%%%%%%%%%%%%%%%%%%%%%%%%%%%%%%%%%%%%%%%%%%%%%%%%%%%%%%%%%%%%%%%%%%%%%%%%
\begin{acks}
	This work was funded by the WALK-MAN project (European Community's 7th Framework Programme: FP7-ICT 611832).
\end{acks}

\bibliographystyle{SageH}
\bibliography{Biblio}

\end{document}